\documentclass[10pt,twocolumn,letterpaper]{article}

\usepackage{cvpr}              
\usepackage{url}            
\usepackage{booktabs}       
\usepackage{amsfonts}       
\usepackage{nicefrac}       
\usepackage{microtype}      
\usepackage{xcolor}         
\usepackage{graphicx}
\usepackage{tcolorbox}
\usepackage{enumitem}
\usepackage{caption}
\usepackage{multirow}
\usepackage{colortbl}
\usepackage{amsmath}
\usepackage{pifont}
\usepackage{algorithm}
\usepackage{algpseudocode}
\usepackage{amsmath}
\usepackage{multicol}

\definecolor{myblue}{RGB}{210, 225, 255}
\usepackage[table]{xcolor}
\usepackage{multirow}
\usepackage{xcolor}
\usepackage{pifont}
\usepackage{booktabs}
\newcommand{\inc}[1]{\textcolor{green!60!black}{\scriptsize(+#1)}}
\newcommand{\dec}[1]{\textcolor{red!70!black}{\scriptsize(-#1)}}
\newcommand{\cmark}{{\color{green!60!black}\ding{51}}} 
\newcommand{\xmark}{{\color{red!70!black}\ding{55}}} 

\definecolor{cvprblue}{rgb}{0.21,0.49,0.74}
\usepackage[pagebackref,breaklinks,colorlinks,allcolors=cvprblue]{hyperref}

\def\paperID{13344} 
\def\confName{CVPR}
\def\confYear{2026}

\title{OpenSubject: Leveraging Video-Derived Identity and Diversity Priors for Subject-driven Image Generation and Manipulation}

\author{
Yexin Liu\textsuperscript{1,2} \quad
Manyuan Zhang\textsuperscript{2,*} \quad
Yueze Wang\textsuperscript{3} \quad
Hongyu Li\textsuperscript{2} \quad
Dian Zheng\textsuperscript{2} \quad
Weiming Zhang\textsuperscript{4} \quad\\
Changsheng Lu\textsuperscript{1} \quad
Yan Feng\textsuperscript{2} \quad
Peng Pei\textsuperscript{2} \quad
Xunliang Cai\textsuperscript{2} \quad
Harry Yang\textsuperscript{1$\dagger$} \\
\\
$^1$HKUST \quad
$^2$Meituan \quad
$^3$Independent Researcher \quad
$^4$HKUST(GZ) \quad\\
\noindent\textsuperscript{*}Project Leader \quad
\textsuperscript{$\dagger$}Corresponding Author
\\
{yliu292@connect.ust.hk, harryyang.hk@gmail.com} \\
GitHub Link: https://github.com/LAW1223/OpenSubject}

\begin{document}
\twocolumn[{
\renewcommand\twocolumn[1][t!]{#1}%
\maketitle
\begin{center}
    \centering
    \includegraphics[width=\textwidth]{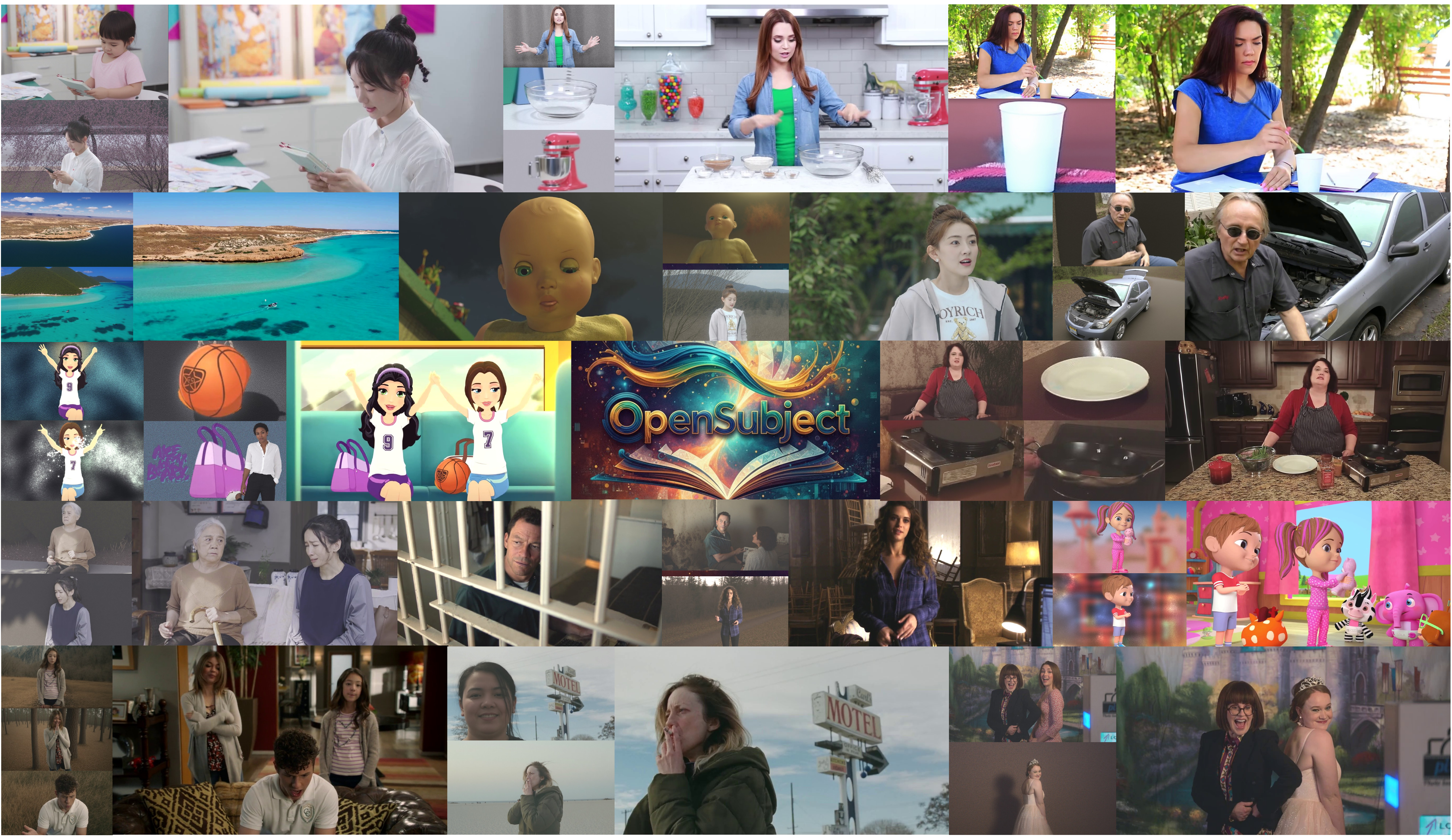}
    \vspace{-20pt}
    \captionof{figure}{\textit{OpenSubject} examples illustrating single- and multi-subject driven image generation and manipulation across human, object, and cartoon domains, spanning indoor and outdoor scenes and diverse viewpoints, and highlighting identity fidelity and contextual diversity.}
    \label{cover_figure}
    \vspace{-5pt}
\end{center}}]

\begin{abstract}
Despite the promising progress in subject-driven image generation, current models often deviate from the reference identities and struggle in complex scenes with multiple subjects. To address this challenge, we introduce \textbf{OpenSubject}, a video-derived large-scale corpus with 2.5M samples and 4.35M images for subject-driven generation and manipulation. The dataset is built with a four-stage pipeline that exploits cross-frame identity priors. (i) \textbf{Video Curation}. We apply resolution and aesthetic filtering to obtain high-quality clips. (ii) \textbf{Cross-Frame Subject Mining and Pairing}.  We utilize vision-language model (VLM)–based category consensus, local grounding, and diversity-aware pairing to select image pairs. (iii) \textbf{Identity-Preserving Reference Image Synthesis}. We introduce segmentation map-guided outpainting to synthesize the input images for subject-driven generation and box-guided inpainting to generate input images for subject-driven manipulation, together with geometry-aware augmentations and irregular boundary erosion. (iv) \textbf{Verification and Captioning}. We utilize a VLM to validate synthesized samples, re-synthesize failed samples based on stage (iii), and then construct short and long captions. In addition, we introduce a benchmark covering subject-driven generation and manipulation, and then evaluate identity fidelity, prompt adherence, manipulation consistency, and background consistency with a VLM judge. Extensive experiments show that training with \textbf{OpenSubject} improves generation and manipulation performance, particularly in complex scenes. The project is available on \href{https://github.com/LAW1223/OpenSubject}{GitHub}.
\end{abstract}    
\section{Introduction}
\label{sec:intro}

\begin{table*}[t]
\centering
\caption{Comparison of subject-driven image generation and manipulation datasets. Counts denote the number of \emph{paired input samples}.}
\label{tab_dataset_comparison}
\vspace{-6pt}
\setlength{\tabcolsep}{8pt}
\renewcommand{\arraystretch}{1.08}
\resizebox{\textwidth}{!}{\begin{tabular}{l|c|c|c|c|c|c}
\toprule
\multirow{2}{*}{\textbf{Dataset}} & \textbf{Paired Single-subject} & \textbf{Paired Multi-subject} & \textbf{Subject-driven} & \multirow{2}{*}{\textbf{IP Source}} & \textbf{General} & \textbf{Diverse} \\
 & \textbf{Input Samples} & \textbf{Input Samples} & \textbf{Manipulation} &  & \textbf{Objects} & \textbf{Contexts} \\
\midrule
Echo-4o-Image~\cite{ye2025echo}      & \xmark & 73k    & \xmark & Synthesis & \cmark & \xmark \\
UNO-1M~\cite{uno}             & 1M     & \xmark & \xmark & Synthesis & \cmark & \xmark \\
Subjects200K~\cite{xie2023omnicontrol}        & 200k   & \xmark & \xmark & Synthesis & \cmark & \xmark \\
MultiID-2M~\cite{xu2025withanyone}         &  \xmark  & 500k & \xmark & Retrieval & \xmark & \xmark \\
\midrule
\textbf{OpenSubject (ours)} & \textbf{748k} & \textbf{1752k} & \textbf{\cmark} & \textbf{Video} & \textbf{\cmark} & \textbf{\cmark} \\
\bottomrule
\end{tabular}}
\vspace{-12pt}
\end{table*}

Subject-driven image generation aims to synthesize realistic images of specified subjects conditioned on text prompts while preserving identity~\cite{gal2022image,ruiz2023dreambooth}. This capability underpins applications in personalized content creation~\cite{mou2025dreamo}, portrait or character re-rendering~\cite{jiang2025infiniteyou}, and interactive editing~\cite{wang2024ms,huang2025resolving}. With the advent of Diffusion Transformers (DiTs)~\cite{peebles2023scalable}, methods in this area have achieved substantial progress, particularly in the single-subject setting, enabling identity-consistent image generation. 

Recent work has shifted toward multi-subject personalization~\cite{uno,chen2025xverse,cheng2025umo}, in which a model composes several subjects into a single, coherent scene while preserving each identity. Compared with the single-subject setting, this scenario is considerably more challenging: the model must maintain per-subject identity under novel poses and contexts and follow the textual specification.

To address these issues, prior work has pursued two paradigms for constructing paired samples, synthesis-based and retrieval-based, as summarized in Tab.~\ref{tab_dataset_comparison}. Synthesis-based pipelines~\cite{ye2025echo,uno,xie2023omnicontrol} typically leverage text-to-image (T2I) or image-to-image models to produce paired exemplars with controlled variation but often inherit model biases and exhibit identity inconsistency. Retrieval-based approaches~\cite{xu2025withanyone} collect web images (e.g., using celebrity names), but struggle to scale to multi-subject settings, are skewed toward public figures, and provide limited coverage of general object categories.

In this work, we leverage video as an underexplored yet promising source of identity-consistent supervision. Videos naturally provide multi-frame observations of subjects with rich variation in viewpoint, illumination, and environment, making them well-suited for learning appearance priors for multi-reference conditioning. However, two challenges remain: 1) selecting cross-frame pairs that maximize diversity while preserving identity consistency; and 2) ensuring sufficient context variation, as naive within-clip pairing often yields near-identical backgrounds and thus weak supervision for subject-driven generation.

To address these challenges, we introduce \textbf{OpenSubject}, a dataset constructed via a four-stage pipeline. (i) \textit{Video curation}: we collect large-scale open-source videos and apply resolution and aesthetic filters to obtain high-quality, subject-persistent clips. (ii) \textit{Cross-frame subject mining and pairing}: we sample candidate frames, enforce clip-level subject consensus with a vision–language model (VLM), perform instance-level local verification with Grounding-DINO~\cite{liu2024grounding} and the VLM, and select a maximally diverse frame pair using DINOv2 embeddings~\cite{oquab2023dinov2}. (iii) \textit{Identity-preserving reference image synthesis}: we use mask-guided outpainting to construct cross-frame inputs for subject-driven generation, complemented by geometry-aware augmentations and irregular boundary erosion; we also introduce an editing task, \textit{subject-driven manipulation}, defined as replacing the target subject in the input image with the referenced identity while preserving all non-target content, for which we synthesize inputs via box-guided inpainting. (iv) \textit{Verification and captioning}: we validate synthesized samples with a VLM, re-synthesize failures based on stage (iii), and generate both short and long captions. The resulting corpus (2.5M samples, 4.35M images) supports both subject-driven generation and manipulation (see Fig.~\ref{cover_figure}). OpenSubject is a video-grounded yet synthetic-paired corpus: real videos define the subject identities, poses, and multi-view context, while FLUX.1 Fill [dev]~\cite{flux1-fill-dev} in-/out-painting is used to synthesize the inputs.

We further establish an open-ended benchmark that spans subject-driven generation and manipulation, comprising four sub-tasks and evaluated with a rubricized VLM judge (GPT-4.1~\cite{openai_gpt4_1_2025}), reporting identity fidelity, prompt adherence, manipulation fidelity, and background consistency. Extensive experiments demonstrate that models trained on OpenSubject achieve stronger identity preservation and manipulation fidelity, particularly in multi-subject settings.

Our main contributions are as follows: (i) \textbf{Large-scale, video-derived corpus for subject-driven generation and manipulation}. We present OpenSubject, a 2.5M-sample (4.35M-image) dataset for subject-driven image generation and manipulation, covering both single- and multi-subject settings. (ii) \textbf{Scalable video-based construction pipeline}. We develop a four-stage pipeline that ensures subject consistency, diversity, and scalability. (iii) \textbf{Open-ended benchmark and evaluation protocol}. We introduce a benchmark with four sub-tasks and a rubricized VLM judge, and demonstrate that training on OpenSubject yields improvements on both our benchmark and external suites.
\section{Related Works}
\label{sec: Related Works}

\noindent \textbf{Subject-driven image generation.} Subject-driven image generation aims to synthesize identity-consistent images of specified subjects. Existing approaches fall into two architectural families: U-Net–based and DiT-based. Early work is predominantly U-Net–based; for example, DreamBooth~\cite{ruiz2023dreambooth} adapts Stable Diffusion~\cite{rombach2022high} for single-subject personalization via test-time fine-tuning~\cite{gal2022image,ruiz2024hyperdreambooth,kumari2023multi}. Zero-shot variants such as IP-Adapter~\cite{ye2023ip}, PhotoVerse~\cite{chen2023photoverse}, FastComposer~\cite{xiao2025fastcomposer}, and FlashFace~\cite{zhang2024flashface} improve inference efficiency by injecting identity features through vision encoders or lightweight adapters, yet they inherit limitations of U-Net architectures. By contrast, recent DiT-based methods offer stronger representational capacity and more flexible conditioning. For single-subject generation, models such as IC-LoRA~\cite{huang2024context}, InstantID~\cite{wang2024instantid}, and InfiniteYou~\cite{jiang2025infiniteyou} introduce mechanisms including residual identity injection and attention-based control to improve identity fidelity and prompt alignment. Extending to multi-subject generation, frameworks including UNO~\cite{xie2023omnicontrol}, OmniControl~\cite{xie2023omnicontrol}, UniReal~\cite{chen2025unireal}, DreamO~\cite{mou2025dreamo}, XVerse~\cite{chen2025xverse}, OmniGen~\cite{xiao2025omnigen,wu2025omnigen2}, and Emu3.5~\cite{cui2025emu3} employ multi-image conditioning, token-level constraints, and text-stream modulation to enable personalized control over multiple identities. Despite this progress, achieving scalable, high-quality multi-subject generation, especially in open-ended scenes, remains a significant challenge. To mitigate this challenge, we introduce OpenSubject, a large-scale corpus for subject-driven generation and manipulation.

\noindent \textbf{Datasets for subject-driven image generation.} High-quality datasets are essential for preserving identity across diverse poses, illumination, and scenes. Prior work follows two paradigms: synthesis-based and retrieval-based. Synthesis-based corpora, such as Subjects200K~\cite{xie2023omnicontrol} and UNO-1M~\cite{xie2023omnicontrol}, use strong base models (e.g., FLUX.1~\cite{flux1controlnet}) with prompts to create identity-consistent pairs across varied contexts; Echo-4o-Image~\cite{ye2025echo} extends this direction with roughly 180K long-tailed samples that complement real-world coverage. Retrieval-based datasets, exemplified by MultiID-2M~\cite{xu2025withanyone}, curate real-world group photos paired with individual identity references. However, these resources largely rely on static web images or synthetic pairings and therefore lack scalable multi-reference coverage and frame-level identity coherence. We address these gaps with OpenSubject, a million-scale, video-derived dataset that captures cross-frame identity priors and supports both subject-driven generation and subject-driven manipulation. By leveraging temporal consistency and controlled reference synthesis, OpenSubject improves prompt adherence and multi-reference fidelity in subject-conditioned models.
\section{OpenSubject}
\label{OpenSubject}

\begin{figure*}[!t]
\centering
\includegraphics[width=\linewidth]{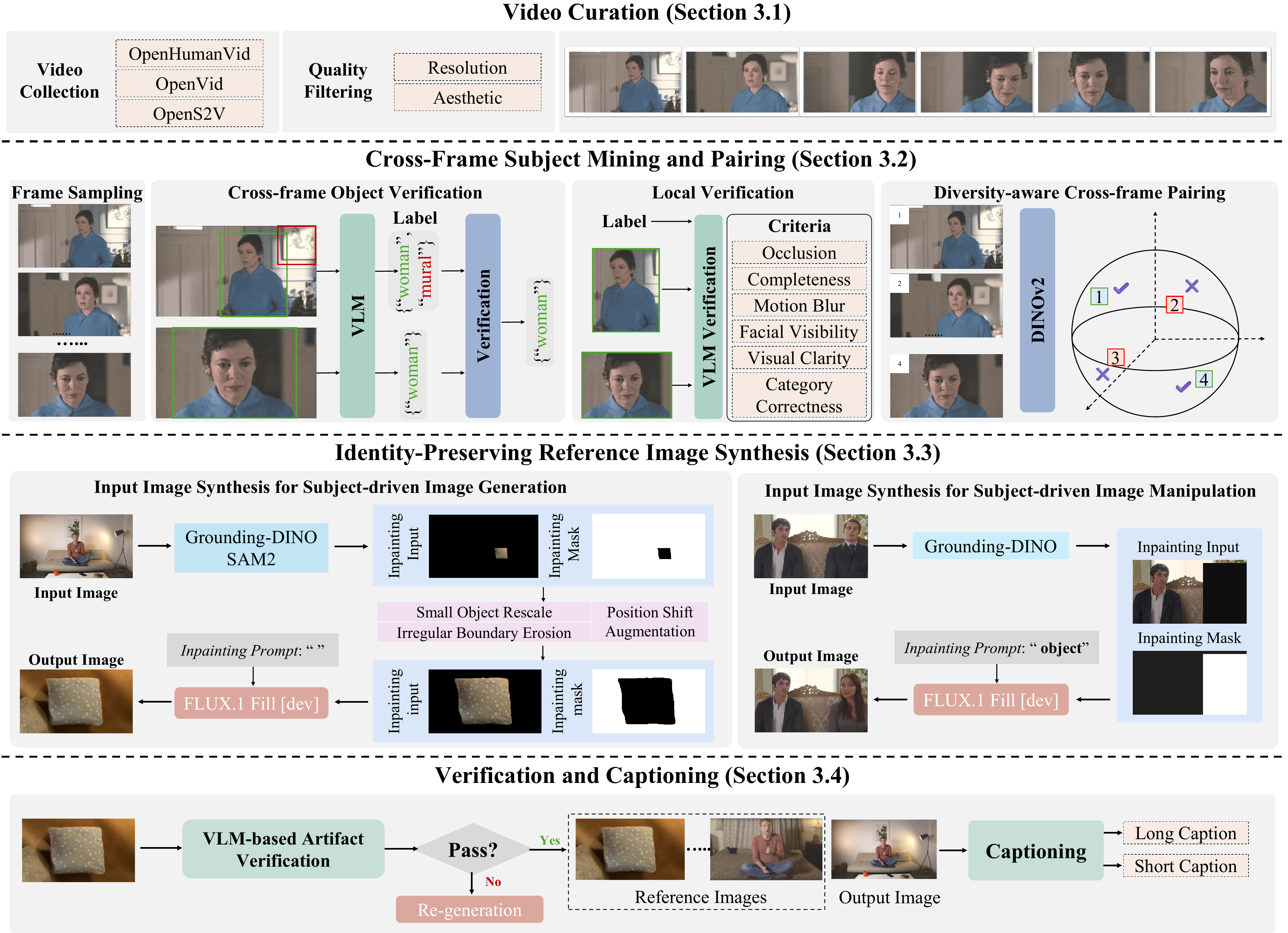}
\vspace{-8pt}
\caption{Overview of the \textbf{OpenSubject} pipeline. (a) \textbf{Video curation}: collect videos from OpenHumanVid, OpenVid, and OpenS2V, and apply resolution and aesthetic filters. (b) \textbf{Cross-frame subject mining and pairing}: verify objects with a vision–language model (category consensus, visual clarity, occlusion, facial visibility), localize with Grounding-DINO, and select diverse frame pairs. (c) \textbf{Identity-preserving reference synthesis}: use mask-guided outpainting for generation and box-guided inpainting for manipulation. (d) \textbf{Automated verification and captioning}: perform VLM-based artifact checks and regenerate failures, then produce short and long captions for training.}
\vspace{-12pt}
\label{framework}
\end{figure*}
Fig.~\ref{framework} provides an overview of the \textbf{OpenSubject} data-construction pipeline, which comprises video curation, cross-frame subject mining and pairing, identity-preserving reference image synthesis, and verification and captioning.

\subsection{Video Curation}
\noindent\textbf{Video collection.} We construct OpenSubject from large-scale, publicly available video corpora (OpenVid~\cite{nan2024openvid}, OpenHumanVid~\cite{li2025openhumanvid}, OpenS2V~\cite{yuan2025opens2v}).

\noindent\textbf{Quality filtering.} We discard videos that do not meet quality requirements, including those with insufficient spatial resolution (below 720p) or low aesthetic scores (below 5.8).

\begin{figure*}[!t]
\centering
\includegraphics[width=\linewidth]{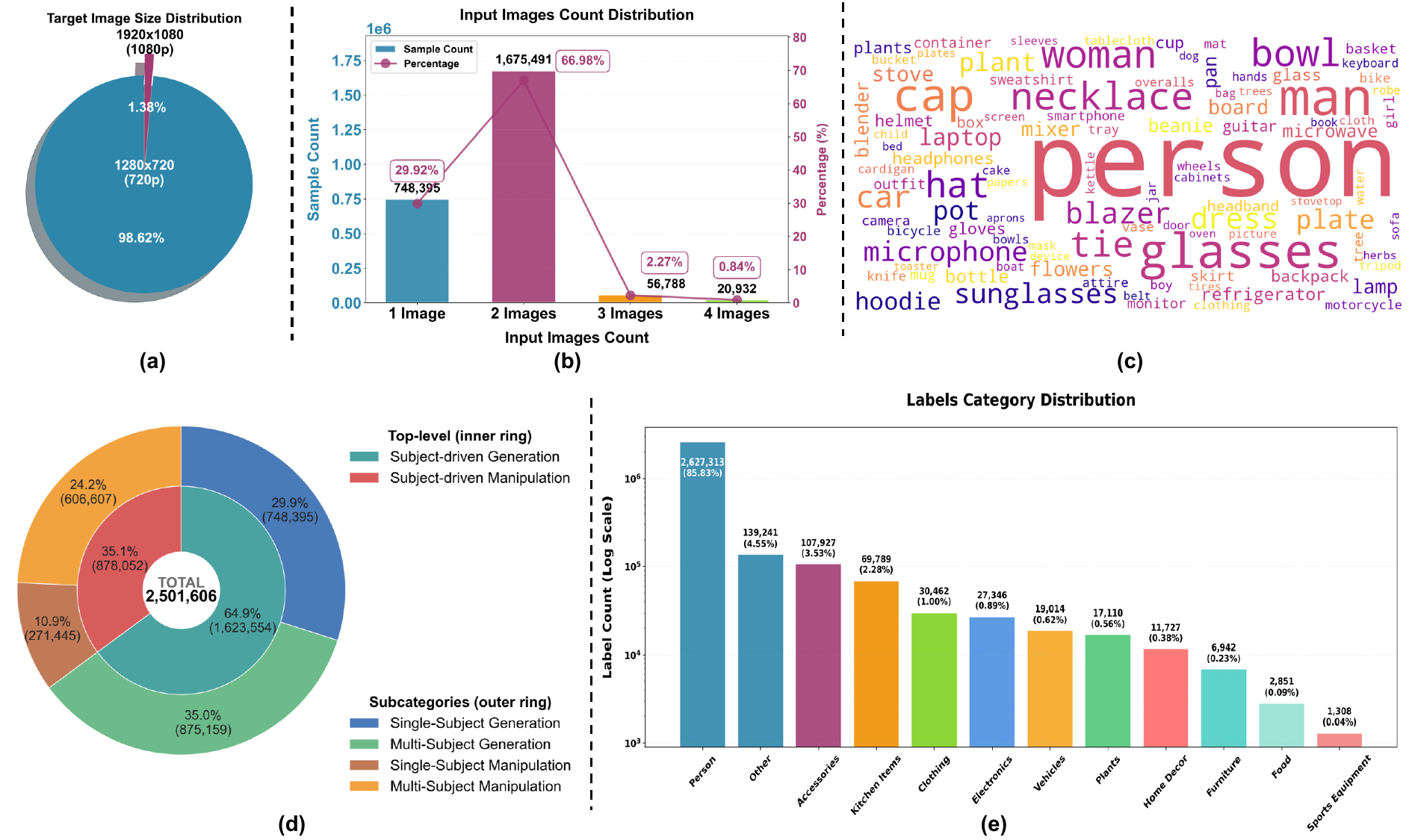}
\vspace{-15pt}
\caption{\textbf{Dataset statistics of \textbf{OpenSubject}.} 
(a) Spatial resolution distributions. 
(b) Distribution of the number of references per sample. 
(c) Word cloud for subjects.
(d) Task composition across four sub-tasks.
(e) Subject category frequency. } 
\vspace{-10pt}
\label{statistics}
\end{figure*}

\subsection{Cross-Frame Subject Mining and Pairing}

To construct reliable cross-frame training pairs, we use a hierarchical procedure: (i) uniformly sample four mid-range frames; (ii) enforce clip-level subject consensus, retaining only frames that share the same detected subject category; (iii) perform instance-level verification via grounding and a VLM to remove spurious detections; and (iv) select a diversity-aware pair by choosing the two filtered frames with maximal DINOv2~\cite{oquab2023dinov2} embedding distance.

\noindent \textbf{Frame sampling.} We uniformly sample four mid-range frames to capture natural appearance variation without redundancy, yielding candidates that differ meaningfully in viewpoint and context.

\noindent \textbf{Cross-frame subject verification.} We apply Qwen2.5-VL-7B~\cite{bai2025qwen2} to each candidate frame to detect foreground subjects and retain only instances occupying at least 5\% of the image area. We then enforce cross-frame consensus by computing the intersection (or majority agreement) of subject categories across frames and keeping only frames/instances whose category belongs to this consensus set. If the set is empty, the clip is discarded.

\noindent \textbf{Local verification.} To ensure spatial and semantic validity, we run instance-level grounding with Grounding-DINO~\cite{liu2024grounding} using class-specific confidence thresholds (0.5 for objects, 0.8 for humans). Detections are filtered using geometric priors—box count, relative area, aspect ratio, and pairwise IoU—to remove spurious or degenerate cases. The corresponding regions of interest are then cropped and evaluated by Qwen2.5-VL-7B~\cite{bai2025qwen2} for (i) label correctness, (ii) occlusion, (iii) completeness, (iv) motion blur, and (v) facial visibility (for human subjects).

\noindent \textbf{Diversity-aware cross-frame pairing.} To broaden scene-context coverage, we compute DINOv2's embeddings for the filtered frames and select a pair with the largest cosine distance. Clips with fewer than two frames are discarded.

\subsection{Identity-Preserving Reference Image Synthesis}
Given curated frames, we define two subject-conditioned tasks: subject-driven generation (input and target drawn from different frames of the same clip) and subject-driven manipulation (edits within a single frame). Ground-truth targets are the original video frames, while inputs are synthesized by mask-guided outpainting (for generation) or inpainting (for manipulation). To standardize resolution and improve visual quality, we apply geometry-aware augmentations (scale and position jitter) and perform irregular boundary erosion on masks to reduce seam artifacts (e.g., banding, black borders) before synthesis.

\noindent \textbf{Input image synthesis for subject-driven generation.}
Given verified subject labels and bounding boxes, we synthesize training inputs via mask-guided outpainting.
(i) \emph{Fine mask construction.} We refine localization with Grounding-DINO and SAM2~\cite{ravi2024sam} to obtain instance masks.
(ii) \emph{Mask topology and geometry normalization.} To mitigate identity leakage, overlapping masks are reconciled by subtracting nested regions from larger instances. We then apply geometry augmentations (scale and placement): instances with area \(<30\%\) are upscaled to a random target in \(30\text{--}40\%\) (aspect ratio preserved), and all subjects are re-centered with bounded jitter to enhance layout diversity while maintaining identity fidelity.
(iii) \emph{Outpainting.} Using subject pixels and refined masks as constraints, we perform outpainting with FLUX.1 Fill [dev]~\cite{flux1-fill-dev} to complete the surrounding context, yielding identity-preserving, composition-ready inputs for subject-driven generation.

\noindent \textbf{Input image synthesis for subject-driven manipulation.}
We construct input images for the manipulation task via mask-guided inpainting.
(i) \emph{Sample selection.} We choose multi-object images with low pairwise box overlap and randomly select one or more target instances as editing regions.
(ii) \emph{Mask construction.} Unlike the generation branch, segmentation masks are not required. Instead, the bounding box is used to define the erase mask and the corresponding input conditioning.
(iii) \emph{Inpainting.} FLUX.1 Fill [dev] is applied to complete the erased regions. The resulting pairs provide precise localization supervision for referent replacement while preserving non-target content.

\subsection{Verification and Captioning}
We adopt a verify–refine–caption stage. Qwen2.5-VL-7B acts as the verifier, assessing synthesized samples for artifacts and physical plausibility. Samples that fail are re-synthesized with a different random seed. For accepted samples, Qwen2.5-VL-7B then generates two caption variants (short and long), with the instruction style randomly instantiated as either generation or editing.

\subsection{Statistical Analysis}
We construct the \textbf{OpenSubject} dataset with the above pipeline, yielding 2.5M samples and 4.35M images. As shown in Fig.~\ref{statistics} (a–e), resolutions are predominantly 720p, with a smaller 1080p subset. The reference set includes a single-image subset (748k), while the remaining samples contain two or more input images, enabling both single- and multi-reference settings. The corpus spans four sub-tasks: single-subject generation, single-subject manipulation, multi-subject generation, and multi-subject manipulation, with generation accounting for roughly two-thirds of the data. Category and scene coverage is broad, encompassing people, objects, and diverse environments.

\begin{table*}[t]
\centering
\caption{Quantitative results of single-subject and multi-subject driven generation and manipulation on the OSBench.}
\label{tab:xversebench}
\vspace{-5pt}
\setlength{\tabcolsep}{3pt}   
\renewcommand{\arraystretch}{}
\resizebox{\textwidth}{!}{%
\begin{tabular}{l|c|ccc|ccc|ccc|ccc|c}
\toprule
\multirow{3}{*}{Method} & \multirow{3}{*}{Size (B)} &
\multicolumn{6}{c|}{Subject-driven generation} &
\multicolumn{6}{c|}{Subject-driven manipulation} &
\multirow{3}{*}{Average $\uparrow$} \\
\cmidrule(lr){3-8} \cmidrule(lr){9-14}
& &\multicolumn{3}{c|}{Single-Subject} & \multicolumn{3}{c|}{Multi-Subject} & \multicolumn{3}{c|}{Single-Subject} & \multicolumn{3}{c|}{Multi-Subject}\\
\cmidrule(lr){3-8} \cmidrule(lr){9-14}
&& PA & IF & Overall & PA & IF & Overall & MF & BC & Overall & MF & BC & Overall\\
\midrule
\rowcolor[HTML]{EFEFEF}
\multicolumn{15}{c}{\textit{Closed-source models}}\\
\midrule
Gemini 2.5 Flash Image Preview~\citep{google2025gemini25flashmodelcard} & - & 9.07 & 8.83 & 8.92 & 8.17 & 8.42 & 8.14 & 7.23 & 8.20 & 7.16 & 6.90 & 4.87 & 5.12 & 7.34 \\
GPT-4o-2024-11-20~\cite{gpt4o} & - &9.23 & 8.49 & 8.82 & 8.39 & 7.59 & 7.82 & 7.64 & 6.22 & 6.80 & 7.10 & 2.93 & 3.77 & 6.80\\
Gemini 2.0 Flash Image Preview~\cite{gemini-2.0-flash} & - & 8.33 & 8.30 & 8.27 & 7.62 & 7.38 & 7.39 & 3.37 & 3.22 & 2.25 & 4.98 & 2.38 & 2.81 & 5.18\\
\hline
\rowcolor[HTML]{EFEFEF}
\multicolumn{15}{c}{\textit{Open-source models}}\\
\hline
UNO~\cite{uno} & 12 &7.53 & 5.30 & 5.95 & 7.13 & 6.12 & 6.40 & 2.83 & 0.55 & 0.78 & 1.45 & 0.85 & 0.73 & 3.46\\
DreamO~\cite{mou2025dreamo} & 12 & 7.10 & 4.12 & 5.13 & 7.03 & 6.18 & 6.52 & 3.86 & 0.27 & 0.38 & 2.13 & 0.71 & 0.72 & 3.19 \\
XVerse~\cite{chen2025xverse} & 12 &6.58 & 1.95 & 3.11 & 7.00 & 5.28 & 5.95 & 3.15 & 0.38 & 0.65 & 0.82 & 0.67 & 0.38 & 2.52\\
DreamOmni2~\cite{xia2025dreamomni2} & 19 & 8.00 & 7.35 & 7.53 & 7.42 & 6.62 & 6.88 & 6.08 & 6.30 & 5.72 & 5.10 & 1.40 & 2.16 & 5.57\\
Qwen-Image-Edit-2509~\citep{wu2025qwen} & 20 &9.03 & 8.55 & 8.77 & 7.83 & 7.25 & 7.44 & 7.63 & 6.53 & 6.69 & 7.90 & 3.85 & 5.10 & 7.00\\
OmniGen2~\citep{wu2025omnigen2} & 7 &8.50 & 8.70 & 8.55 & 7.98 & 7.97 & 7.89 & 5.92 & 5.00 & 4.99 & 7.37 & 3.27 & 4.31 & 6.43\\
\bottomrule
\end{tabular}%
}
\vspace{-2pt}
\end{table*}

\begin{table*}[t]
\centering
\caption{Ablation study of fine-tuning with OpenSubject. Row 2 adds synthetic T2I data to improve prompt adherence. Row 3 additionally incorporates sampled OpenSubject data on top of T2I. \textcolor{green!60!black}{Green} and \textcolor{red!70!black}{red} indicate performance improvements and decreases, respectively.}
\label{tab:ablation}
\vspace{-6pt}
\setlength{\tabcolsep}{3.2pt}
\renewcommand{\arraystretch}{1}
\resizebox{\textwidth}{!}{%
\begin{tabular}{l|ccc|ccc|ccc|ccc|c}
\toprule
\multirow{3}{*}{Method} &
\multicolumn{6}{c|}{\textbf{Subject-driven generation}} &
\multicolumn{6}{c|}{\textbf{Subject-driven manipulation}} &
\multirow{3}{*}{Average$\uparrow$} \\
\cmidrule(lr){2-7}\cmidrule(lr){8-13}
& \multicolumn{3}{c|}{Single-Subject} & \multicolumn{3}{c|}{Multi-Subject} & \multicolumn{3}{c|}{Single-Subject} & \multicolumn{3}{c|}{Multi-Subject} \\
\cmidrule(lr){2-7}\cmidrule(lr){8-13}
& PA & IF & Overall & PA & IF & Overall & MF & BC & Overall & MF & BC & Overall \\
\midrule
\textbf{OmniGen2 (baseline)} 
& 8.50 & 8.70 & 8.55 & 7.98 & 7.97 & 7.89 & 5.92 & 5.00 & 4.99 & 7.37 & 3.27 & 4.31 & 6.43 \\
\midrule
+ T2I 
& 8.62 \inc{0.12} & 8.17 \dec{0.53} & 8.33 \dec{0.22}
& 8.12 \inc{0.14} & 7.72 \dec{0.25} & 7.81 \dec{0.08}
& 5.08 \dec{0.84} & 3.22 \dec{1.78} & 3.60 \dec{1.39}
& 6.47 \dec{0.90} & 1.97 \dec{1.30} & 2.98 \dec{1.33}
& 5.68 \dec{0.75} \\
\rowcolor{cyan!5} + \textbf{OpenSubject (ours)} 
& 8.30 \dec{0.20} & 8.95 \inc{0.25} & 8.58 \inc{0.03}
& 8.05 \inc{0.07} & 8.65 \inc{0.68} & 8.26 \inc{0.37}
& 7.18 \inc{1.26} & 5.22 \inc{0.22} & 5.80 \inc{0.81}
& 7.85 \inc{0.48} & 5.20 \inc{1.93} & 6.22 \inc{1.91}
& 7.22 \inc{0.79} \\
\bottomrule
\end{tabular}%
}
\vspace{-8pt}
\end{table*}
\section{Benchmark}
\label{benchmark}

Prior benchmarks (e.g., XVerse~\cite{chen2025xverse}) focus on clean, single-subject portraits and rarely evaluate subjects in complex scenes; moreover, subject-driven manipulation is typically omitted. We introduce a benchmark (named OSBench) that spans subject-driven generation and manipulation tasks.

\noindent\textbf{Tasks.} The benchmark comprises four sub-tasks (60 samples each): (i) \textbf{Single-subject generation}. Synthesize an identity-consistent image from one reference under an open-ended prompt; (ii) \textbf{Multi-subject generation}. Synthesize an image by fusing two to four references under an open-ended prompt; (iii) \textbf{Single-subject manipulation}. Replace one target in a scene that contains a single principal object; and (iv) \textbf{Multi-subject manipulation}. Replace one target in a complex scene (containing multiple subjects) while preserving non-target content.

\noindent\textbf{Evaluation dimensions.} Following instruction-based assessment methods (e.g., VIEScore~\cite{ku2023viescore}, OmniContext~\cite{wu2025omnigen2}), we use a strong VLM judge (GPT-4.1~\cite{openai_gpt4_1_2025}) to assign 0–10 scores with rubricized prompts and independent criteria. For \textbf{generation} tasks, we report Prompt Adherence (PA) (attribute/count/relation compliance), Identity Fidelity (IF) (consistency with the subject across provided references), and Overall (geometric mean of PA and IF). For \textbf{manipulation} tasks, we report Manipulation Fidelity (MF) (match between edited regions and the referenced subject(s)), Background Consistency (BC) (stability of non-edited regions), and Overall (geometric mean of MF and BC).

\begin{table*}[t]
\centering
\caption{Quantitative comparison results on OmniContext. ``Char. + Obj." indicates Character + Object. Fine-tuning on \textbf{OpenSubject} yields large gains on the MULTIPLE subset and consistent improvements on SINGLE and SCENE.}
\vspace{-6pt}
\renewcommand{\arraystretch}{0.95}
\resizebox{0.99\linewidth}{!}{
\begin{tabular}{l|c|cc|ccc|ccc|c}
\toprule
\multirow{2}{*}{\bf Model} & 
\multirow{2}{*}{\bf Size (B)} & 
\multicolumn{2}{c|}{\bf SINGLE} & 
\multicolumn{3}{c|}{\bf MULTIPLE} & 
\multicolumn{3}{c|}{\bf SCENE} & 
\multirow{2}{*}{\bf Average$\uparrow$} \\
\cmidrule(lr){3-10}
& & Character & Object & Character & Object & Char. + Obj. & Character & Object & Char. + Obj. & \\
\midrule
\rowcolor[HTML]{EFEFEF}
\multicolumn{11}{c}{\textit{Closed-source models}}\\
\midrule
FLUX.1 Kontext [Max]~\cite{labs2025flux} & - & 8.48 & 8.68 & - & - & - & - & - & - & - \\
Gemini 2.5 Flash Image Preview~\citep{google2025gemini25flashmodelcard} & - & 8.52 & 9.14 & 7.80 & 8.64 & 6.63 & 6.74 & 7.11 & 6.04 & 7.58 \\
Gemini 2.5 Flash Image~\citep{google2025gemini25flashmodelcard} & - & 8.62 & 8.91 & 7.88 & 8.92 & 7.39 & 7.29 & 7.05 & 6.68 & 7.84 \\
GPT-4o~\cite{gpt4o} & - & 8.90 & 9.01 & 9.07 & 8.95 & 8.54 & 8.90 & 8.44 & 8.60 & 8.80 \\
Gemini 2.0 Flash~\cite{gemini-2.0-flash} & - & 5.06 & 5.17 & 2.91 & 2.16 & 3.80 & 3.02 & 3.89 & 2.92 & 3.62 \\
\midrule
\rowcolor[HTML]{EFEFEF}
\multicolumn{11}{c}{\textit{Open-source models}}\\
\midrule
Emu3.5~\cite{cui2025emu3} & 32 & 8.72 & 9.46 & 8.65 & 9.09 & 8.78 & 8.78 & 8.89 & 8.15 & 8.82 \\
Qwen-Image-Edit-2509~\citep{wu2025qwen} & 20 & 8.35 & 9.13 & 7.65 & 8.85 & 7.90 & 5.16 & 7.75 & 6.73 & 7.69 \\
OmniGen~\cite{xiao2025omnigen} & 3.8 & 7.21 & 5.71 & 5.65 & 5.44 & 4.68 & 3.59 & 4.32 & 5.12 & 4.34 \\
InfiniteYou~\cite{jiang2025infiniteyou} & 12 & 6.05 & - & - & - & - & - & - & - & - \\
UNO~\cite{uno} & 12 & 6.60 & 6.83 & 2.54 & 6.51 & 4.39 & 2.06 & 4.33 & 4.37 & 4.71 \\
BAGEL~\cite{de2006bagel} & 14 & 5.48 & 7.03 & 5.17 & 6.64 & 6.24 & 4.07 & 5.71 & 5.47 & 5.73 \\
\midrule
Baseline (OmniGen2~\citep{wu2025omnigen2}) & 7 & 8.05 & 7.58 & 7.11 & 7.13 & 7.45 & 6.38 & 6.71 & 7.04 & 7.18 \\
\rowcolor{cyan!5} \textbf{Ours} & 7 
& 8.18 \inc{0.13} 
& 7.54 \dec{0.04} 
& 7.34 \inc{0.23} 
& 7.37 \inc{0.24} 
& 7.87 \inc{0.42} 
& 6.50 \inc{0.12} 
& 6.92 \inc{0.21} 
& 7.00 \dec{0.04} 
& 7.34 \inc{0.16} \\
\bottomrule
\end{tabular}
}
\vspace{-6pt}
\label{tab:omni_context}
\end{table*}

\begin{table*}[!t]
    \centering
    \caption{Quantitative results on ImgEdit~\citep{ye2025imgedit}. Fine-tuning on \textbf{OpenSubject} also yields gains in image editing, especially on the ``Add", ``Extract", ``Background", and ``Hybrid" subsets. The fine-tuned model achieves a higher overall score than FLUX.1 Kontext [Dev]~\citep{labs2025flux}.}
    \label{tab:imgedit}
    \vspace{-6pt}
    \renewcommand{\arraystretch}{0.95}
    \resizebox{1.0\linewidth}{!}{
    \begin{tabular}{l|ccccccccc|c}
        \toprule
        \textbf{Model} & \bf Add & \bf Adjust & \bf Extract & \bf Replace & \bf Remove & \bf Background & \bf Style & \bf Hybrid & \bf Action & \bf Overall $\uparrow$ \\
        \midrule
        \rowcolor[HTML]{EFEFEF}
        \multicolumn{11}{c}{\textit{Closed-source models}}\\
        \midrule
        FLUX.1 Kontext [Pro]~\citep{labs2025flux} & 4.25 & 4.15 & 2.35 & 4.56 & 3.57 & 4.26 & 4.57 & 3.68 & 4.63 & 4.00 
        \\
        GPT-Image-1 [High]~\citep{openai_image_api} & 4.61 & 4.33 & 2.90 & 4.35 & 3.66 & 4.57 & 4.93 & 3.96 & 4.89 & 4.20
        \\
        Gemini 2.5 Flash Image Preview~\citep{google2025gemini25flashmodelcard} & 4.47 & 4.19 & 3.81 & 4.39 & 4.70 & 4.20 & 4.18 & 3.48 & 4.68 & 4.23
        \\
        Gemini 2.5 Flash Image~\citep{google2025gemini25flashmodelcard} & 4.65 & 4.34 & 3.69 & 4.49 & 4.65 & 4.32 & 4.13 & 3.66 & 4.59 & 4.28
        \\
        \midrule
        \rowcolor[HTML]{EFEFEF}
        \multicolumn{11}{c}{\textit{Open-source models}}\\
        \midrule
        AnyEdit~\citep{yu2025anyedit} & 3.18 & 2.95 & 1.88 & 2.47 & 2.23 & 2.24 & 2.85 & 1.56 & 2.65 & 2.45 
        \\    
        UltraEdit~\citep{zhao2024ultraedit} & 3.44 & 2.81 & 2.13 & 2.96 & 1.45 & 2.83 & 3.76 & 1.91 & 2.98 & 2.70 
        \\
        OmniGen~\citep{xiao2025omnigen} & 3.47 & 3.04 & 1.71 & 2.94 & 2.43 & 3.21 & 4.19 & 2.24 & 3.38 & 2.96 
        \\
        ICEdit~\citep{zhang2025context} & 3.58 & 3.39 & 1.73 & 3.15 & 2.93 & 3.08 & 3.84 & 2.04 & 3.68 & 3.05 
        \\
        Step1X-Edit~\citep{liu2025step1x} & 3.88 & 3.14 & 1.76 & 3.40 & 2.41 & 3.16 & 4.63 & 2.64 & 2.52 & 3.06 
        \\
        BAGEL~\citep{de2006bagel} & 3.56 & 3.31 & 1.70 & 3.3 & 2.62 & 3.24 & 4.49 & 2.38 & 4.17 & 3.20 
        \\
        UniWorld-1~\citep{lin2025uniworld} & 3.82 & 3.64 & 2.27 & 3.47 & 3.24 & 2.99 & 4.21 & 2.96 & 2.74 & 3.26 
        \\
        Lego-Edit~\citep{jia2025lego} & 3.67 & 3.82 & 2.47 & 3.22 & 3.39 & 4.47 & 4.01 & 3.18 & 3.24 & 3.50
        \\
        FLUX.1 Kontext [Dev]~\citep{labs2025flux} & 4.12 & 3.80 & 2.04 & 4.22 & 3.09 & 3.97 & 4.51 & 3.35 & 4.25 & 3.71
        \\
        Qwen-Image-Edit~\citep{wu2025qwen} & 4.38 & 4.16 & 3.43 & 4.66 & 4.14 & 4.38 & 4.81 &3.82 & 4.69 & 4.27
        \\
        Qwen-Image-Edit-2509~\citep{wu2025qwen} & 4.32 &  4.36 & 4.04 & 4.64 & 4.52 & 4.37 & 4.84 & 3.39 & 4.71 & 4.35
        \\
        Emu3.5~\cite{cui2025emu3} & 4.61 & 4.32 & 3.96 & 4.84 & 4.58 & 4.35 & 4.79 & 3.69 & 4.57 & 4.41
        \\
        \midrule
        Baseline (OmniGen2~\citep{wu2025omnigen2}) & 3.57 & 3.06 & 1.77 & 3.74 & 3.20 & 3.57 & 4.81 & 2.52 & 4.68 & 3.44 
        \\
        \rowcolor{cyan!5} \textbf{Ours}
        & 4.28 \inc{0.71}
        & 3.27 \inc{0.21}
        & 2.61 \inc{0.84}
        & 3.97 \inc{0.23}
        & 3.43 \inc{0.23}
        & 4.13 \inc{0.56}
        & 4.66 \dec{0.15}
        & 3.28 \inc{0.76}
        & 4.45 \dec{0.23}
        & 3.72 \inc{0.28} \\
        \bottomrule
    \end{tabular}
    }
    \vspace{-15pt}
\end{table*}
\section{Experiment}
\label{Experiment}
\subsection{Experiment Setups}

\begin{figure*}[!t]
\centering
\includegraphics[width=\linewidth]{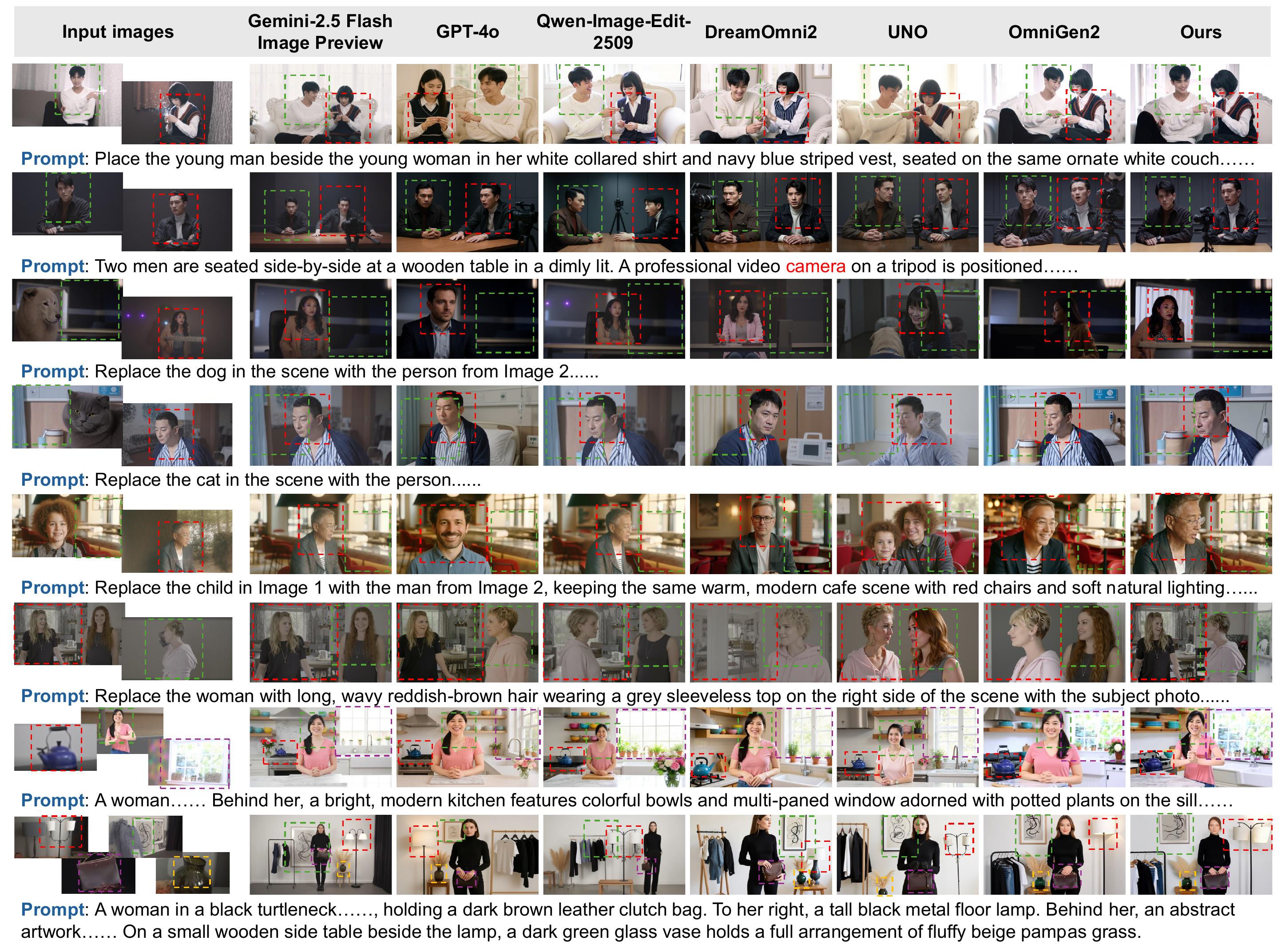}
\vspace{-22pt}
\caption{\textbf{Qualitative comparison.} Colored dashed boxes mark regions of interest for comparison. Boxes of the same color denote corresponding regions across methods and refer to the related area in the input image.} 
\vspace{-15pt}
\label{visualization}
\end{figure*}

\noindent \textbf{Evaluation baselines.} We evaluate both closed- and open-source models. Closed-source methods include Gemini 2.5 Flash Image Preview~\cite{google2025gemini25flashmodelcard}, GPT-4o~\cite{gpt4o}, and Gemini 2.0 Flash Image Preview~\cite{gemini-2.0-flash}. Open-source subject-driven models include UNO~\cite{uno}, DreamO~\cite{mou2025dreamo}, XVerse~\cite{chen2025xverse}, Qwen-Image-Edit-2509~\citep{wu2025qwen}, and OmniGen2~\citep{wu2025omnigen2}.

\noindent \textbf{Implementation details.}
For closed-source models, we use the official inference APIs; when an API does not expose output-resolution control, we specify the target resolution in the prompt.
For open-source models, we use the official checkpoints and default inference settings.
Because Gemini 2.5 Flash Image Preview, Gemini 2.0 Flash Image Preview, and GPT-4o do not permit setting the output resolution, we request the desired resolution via the prompt.
All other models generate one image per item at a fixed resolution.
We also report an open-source baseline fine-tuned on our corpus: OmniGen2 is trained with full-parameter updates on 500k randomly sampled OpenSubject instances plus 100k internal T2I samples to maintain prompt-following ability, using \(16\times\) H800 GPUs.

\subsection{Quantitative Results}
We evaluate on three benchmarks: OSBench, OmniContext~\citep{wu2025omnigen2}, and ImgEdit~\citep{ye2025imgedit}. OSBench and OmniContext assess subject-driven generation, whereas ImgEdit measures instruction-based image editing performance.

\subsubsection{Quantitative evaluation on OSBench.} 

\noindent \textbf{Limitations of existing methods.}
As shown in Tab.~\ref{tab:xversebench}, most approaches struggle on \emph{multi-subject generation} and \emph{subject-driven manipulation}. Open-source models such as UNO, DreamO, and XVerse exhibit very low manipulation performance (single-subject Overall $\leq 0.78$, multi-subject Overall $\leq 0.73$), indicating weak identity replacement and preservation in complex scenes. Strong closed-source models also drop markedly from generation to manipulation. For example, Gemini 2.5 Flash Image Preview attains only 5.12 Overall on multi-subject manipulation despite solid generation scores. Qwen-Image-Edit-2509 is the strongest open-source baseline for manipulation but remains limited in multi-subject cases (Overall 5.10). In summary, current models are not robust when multiple identities must be fused or replaced while keeping non-target content intact.

\noindent \textbf{Fine-tuning with OpenSubject improves quality.}
We evaluate three OmniGen2 configurations: the baseline, the baseline finetuned with 100k T2I samples, and fine-tuned on 500k randomly sampled OpenSubject instances and the T2I data. As shown in Tab.~\ref{tab:ablation}, adding only T2I yields gains in prompt adherence for generation but reduces identity fidelity and weakens manipulation. Fine-tuning on both the OpenSubject and T2I dataset reverses these declines and produces consistent improvements across tasks, raising the Average from 6.43 to 7.22. The largest gains occur in manipulation (single-subject Overall: +0.81; multi-subject Overall: +1.91, with BC: +1.93 and MF: +0.48), while generation benefits in identity fidelity (IF: +0.25 for single-subject and +0.68 for multi-subject) without sacrificing overall performance. These results indicate that OpenSubject provides identity-consistent supervision.

\subsubsection{Quantitative Evaluation on Other Benchmarks}

\textbf{OmniContext}~\citep{wu2025omnigen2}. As shown in Tab.~\ref{tab:omni_context}, fine-tuning OmniGen2 on OpenSubject yields a consistent improvement in average performance from 7.18 to 7.34 (+0.16). Gains concentrate in settings that require integrating multiple references and grounding within scenes: \emph{MULTIPLE/Character} increases from 7.11 to 7.34 (+0.23), \emph{MULTIPLE/Object} from 7.13 to 7.37 (+0.24), and \emph{MULTIPLE/Char.+Obj.} from 7.45 to 7.87 (+0.42). Scene-level categories also improve (\emph{SCENE/Character}: 6.38$\rightarrow$6.50; \emph{SCENE/Object}: 6.71$\rightarrow$6.92), while the single-image case \emph{SINGLE/Object} remains essentially unchanged (7.58$\rightarrow$7.54). These results indicate that OpenSubject primarily enhances identity reasoning and compositional control when multiple entities or strong contextual constraints are present.

\noindent \textbf{ImgEdit}~\citep{ye2025imgedit}. Tab.~\ref{tab:imgedit} shows that \textbf{OpenSubject} fine-tuning improves instruction-based editing for OmniGen2, increasing the overall score from 3.44 to 3.72 (+0.28). The largest gains occur in categories requiring precise localization or structural adjustments: \emph{Extract} (1.77$\rightarrow$2.61, +0.84), \emph{Hybrid} (2.52$\rightarrow$3.28, +0.76), \emph{Add} (3.57$\rightarrow$4.28, +0.71), and \emph{Background} (3.57$\rightarrow$4.13, +0.56). Moderate improvements are observed in \emph{Replace}, \emph{Remove} (+0.23 each), and \emph{Adjust} (+0.21), with minor reductions in \emph{Style} (4.81$\rightarrow$4.66) and \emph{Action} (4.68$\rightarrow$4.45).Overall, these findings indicate that edit robustness has improved across various types of tasks.

\subsection{Qualitative Results}
Fig.~\ref{visualization} presents representative comparisons across methods. Many models struggle to preserve identity in multi-subject scenes and often modify regions outside the intended edits. Compared with OmniGen2, the model finetuned on OpenSubject better maintains subject identity, adheres to attribute specifications, and produces coherent multi-subject compositions. For manipulation, edits remain confined to the marked regions, and non-target content is preserved, whereas other methods frequently exhibit identity drift or unintended changes beyond the edited area.

\section{Conclusion}
We presented \textbf{OpenSubject}, a large-scale, video-derived resource for subject-driven image generation and manipulation, along with a four-stage construction pipeline comprising video curation, cross-frame subject mining and pairing, identity-preserving reference image synthesis, and verification and captioning. We further introduced a benchmark that evaluates single- and multi-subject generation and manipulation using rubricized, aspect-specific criteria. Empirically, prevailing methods degrade on multi-subject generation and subject-conditioned editing in complex scenes. Experiments on our benchmark and external suites demonstrate that fine-tuning on \textbf{OpenSubject} improves identity fidelity and strengthens manipulation robustness.

\noindent \textbf{Ethical considerations.} All data are drawn from publicly available, open-licensed data corpora, and we comply with the original licensing items. The dataset will be released for research-only use under an acceptable-use policy that prohibits biometric identification, re-identification, surveillance, and non-consensual impersonation.

{
    \small
    \bibliographystyle{ieeenat_fullname}
    \bibliography{main}

@String(AAAI = {AAAI})

@article{uno,
  title={Less-to-more generalization: Unlocking more controllability by in-context generation},
  author={Wu, Shaojin and Huang, Mengqi and Wu, Wenxu and Cheng, Yufeng and Ding, Fei and He, Qian},
  journal={arXiv preprint arXiv:2504.02160},
  year={2025}
}

@misc{gemini-2.0-flash,
    author={Google},
    title={Gemini 2.0 Flash},
    year={2025},
    howpublished={\url{https://developers.googleblog.com/en/experiment-with-gemini-20-flash-native-image-generation}},
}

@misc{gpt4o,
    author={OpenAI},
    title={GPT-4o},
    year={2025},
    howpublished={\url{https://openai.com/index/introducing-4o-image-generation}},
}

@article{jiang2025infiniteyou,
  title={InfiniteYou: Flexible photo recrafting while preserving your identity},
  author={Jiang, Liming and Yan, Qing and Jia, Yumin and Liu, Zichuan and Kang, Hao and Lu, Xin},
  journal={arXiv preprint arXiv:2503.16418},
  year={2025}
}

@inproceedings{xiao2025omnigen,
  title={Omnigen: Unified image generation},
  author={Xiao, Shitao and Wang, Yueze and Zhou, Junjie and Yuan, Huaying and Xing, Xingrun and Yan, Ruiran and Li, Chaofan and Wang, Shuting and Huang, Tiejun and Liu, Zheng},
  booktitle={Proceedings of the Computer Vision and Pattern Recognition Conference},
  pages={13294--13304},
  year={2025}
}

@inproceedings{yu2025anyedit,
  title={Anyedit: Mastering unified high-quality image editing for any idea},
  author={Yu, Qifan and Chow, Wei and Yue, Zhongqi and Pan, Kaihang and Wu, Yang and Wan, Xiaoyang and Li, Juncheng and Tang, Siliang and Zhang, Hanwang and Zhuang, Yueting},
  booktitle={Proceedings of the Computer Vision and Pattern Recognition Conference},
  pages={26125--26135},
  year={2025}
}

@article{lin2025uniworld,
  title={Uniworld: High-resolution semantic encoders for unified visual understanding and generation},
  author={Lin, Bin and Li, Zongjian and Cheng, Xinhua and Niu, Yuwei and Ye, Yang and He, Xianyi and Yuan, Shenghai and Yu, Wangbo and Wang, Shaodong and Ge, Yunyang and others},
  journal={arXiv preprint arXiv:2506.03147},
  year={2025}
}

@article{labs2025flux,
  title={FLUX. 1 Kontext: Flow Matching for In-Context Image Generation and Editing in Latent Space},
  author={Labs, Black Forest and Batifol, Stephen and Blattmann, Andreas and Boesel, Frederic and Consul, Saksham and Diagne, Cyril and Dockhorn, Tim and English, Jack and English, Zion and Esser, Patrick and others},
  journal={arXiv preprint arXiv:2506.15742},
  year={2025}
}

@article{wu2025omnigen2,
  title={OmniGen2: Exploration to Advanced Multimodal Generation},
  author={Wu, Chenyuan and Zheng, Pengfei and Yan, Ruiran and Xiao, Shitao and Luo, Xin and Wang, Yueze and Li, Wanli and Jiang, Xiyan and Liu, Yexin and Zhou, Junjie and others},
  journal={arXiv preprint arXiv:2506.18871},
  year={2025}
}

@article{de2006bagel,
  title={BAGEL: a web-based bacteriocin genome mining tool},
  author={de Jong, Anne and van Hijum, Sacha AFT and Bijlsma, Jetta JE and Kok, Jan and Kuipers, Oscar P},
  journal={Nucleic acids research},
  volume={34},
  number={suppl\_2},
  pages={W273--W279},
  year={2006},
  publisher={Oxford University Press}
}

@article{jia2025lego,
  title={Lego-Edit: A General Image Editing Framework with Model-Level Bricks and MLLM Builder},
  author={Jia, Qifei and Liu, Yu and Chai, Yajie and Yao, Xintong and Lu, Qiming and Zhang, Yasen and Shi, Runyu and Huang, Ying and Zhang, Guoquan},
  journal={arXiv preprint arXiv:2509.12883},
  year={2025}
}

@article{liu2025step1x,
  title={Step1x-edit: A practical framework for general image editing},
  author={Liu, Shiyu and Han, Yucheng and Xing, Peng and Yin, Fukun and Wang, Rui and Cheng, Wei and Liao, Jiaqi and Wang, Yingming and Fu, Honghao and Han, Chunrui and others},
  journal={arXiv preprint arXiv:2504.17761},
  year={2025}
}

@article{wu2025qwen,
  title={Qwen-image technical report},
  author={Wu, Chenfei and Li, Jiahao and Zhou, Jingren and Lin, Junyang and Gao, Kaiyuan and Yan, Kun and Yin, Sheng-ming and Bai, Shuai and Xu, Xiao and Chen, Yilei and others},
  journal={arXiv preprint arXiv:2508.02324},
  year={2025}
}

@misc{openai_image_api,
  author       = {{OpenAI}},
  title        = {{Image generation API}},
  howpublished = {\url{https://openai.com/index/image-generation-api/}},
  year={2025}
}

@article{zhang2025context,
  title={In-context edit: Enabling instructional image editing with in-context generation in large scale diffusion transformer},
  author={Zhang, Zechuan and Xie, Ji and Lu, Yu and Yang, Zongxin and Yang, Yi},
  journal={arXiv preprint arXiv:2504.20690},
  year={2025}
}

@misc{google2025gemini25flashmodelcard,
  author       = {Gemini Team},
  title        = {Gemini 2.5 Flash \& Gemini 2.5 Flash Image Model Card},
  howpublished = {\url{https://storage.googleapis.com/deepmind-media/Model-Cards/Gemini-2-5-Flash-Model-Card.pdf}},
  year         = {2025},
}

@article{ye2025imgedit,
  title={Imgedit: A unified image editing dataset and benchmark},
  author={Ye, Yang and He, Xianyi and Li, Zongjian and Lin, Bin and Yuan, Shenghai and Yan, Zhiyuan and Hou, Bohan and Yuan, Li},
  journal={arXiv preprint arXiv:2505.20275},
  year={2025}
}

@article{zhao2024ultraedit,
  title={Ultraedit: Instruction-based fine-grained image editing at scale},
  author={Zhao, Haozhe and Ma, Xiaojian Shawn and Chen, Liang and Si, Shuzheng and Wu, Rujie and An, Kaikai and Yu, Peiyu and Zhang, Minjia and Li, Qing and Chang, Baobao},
  journal={Advances in Neural Information Processing Systems},
  volume={37},
  pages={3058--3093},
  year={2024}
}

@article{wang2024ms,
  title={Ms-diffusion: Multi-subject zero-shot image personalization with layout guidance},
  author={Wang, Xierui and Fu, Siming and Huang, Qihan and He, Wanggui and Jiang, Hao},
  journal={arXiv preprint arXiv:2406.07209},
  year={2024}
}

@article{mou2025dreamo,
  title={Dreamo: A unified framework for image customization},
  author={Mou, Chong and Wu, Yanze and Wu, Wenxu and Guo, Zinan and Zhang, Pengze and Cheng, Yufeng and Luo, Yiming and Ding, Fei and Zhang, Shiwen and Li, Xinghui and others},
  journal={arXiv preprint arXiv:2504.16915},
  year={2025}
}

@article{chen2025xverse,
  title={XVerse: Consistent Multi-Subject Control of Identity and Semantic Attributes via DiT Modulation},
  author={Chen, Bowen and Zhao, Mengyi and Sun, Haomiao and Chen, Li and Wang, Xu and Du, Kang and Wu, Xinglong},
  journal={arXiv preprint arXiv:2506.21416},
  year={2025}
}

@inproceedings{ruiz2023dreambooth,
  title={Dreambooth: Fine tuning text-to-image diffusion models for subject-driven generation},
  author={Ruiz, Nataniel and Li, Yuanzhen and Jampani, Varun and Pritch, Yael and Rubinstein, Michael and Aberman, Kfir},
  booktitle={Proceedings of the IEEE/CVF conference on computer vision and pattern recognition},
  pages={22500--22510},
  year={2023}
}

@article{gal2022image,
  title={An image is worth one word: Personalizing text-to-image generation using textual inversion},
  author={Gal, Rinon and Alaluf, Yuval and Atzmon, Yuval and Patashnik, Or and Bermano, Amit H and Chechik, Gal and Cohen-Or, Daniel},
  journal={arXiv preprint arXiv:2208.01618},
  year={2022}
}

@inproceedings{ruiz2024hyperdreambooth,
  title={Hyperdreambooth: Hypernetworks for fast personalization of text-to-image models},
  author={Ruiz, Nataniel and Li, Yuanzhen and Jampani, Varun and Wei, Wei and Hou, Tingbo and Pritch, Yael and Wadhwa, Neal and Rubinstein, Michael and Aberman, Kfir},
  booktitle={Proceedings of the IEEE/CVF conference on computer vision and pattern recognition},
  pages={6527--6536},
  year={2024}
}

@inproceedings{kumari2023multi,
  title={Multi-concept customization of text-to-image diffusion},
  author={Kumari, Nupur and Zhang, Bingliang and Zhang, Richard and Shechtman, Eli and Zhu, Jun-Yan},
  booktitle={Proceedings of the IEEE/CVF conference on computer vision and pattern recognition},
  pages={1931--1941},
  year={2023}
}

@inproceedings{rombach2022high,
  title={High-resolution image synthesis with latent diffusion models},
  author={Rombach, Robin and Blattmann, Andreas and Lorenz, Dominik and Esser, Patrick and Ommer, Bj{\"o}rn},
  booktitle={Proceedings of the IEEE/CVF conference on computer vision and pattern recognition},
  pages={10684--10695},
  year={2022}
}

@article{ye2023ip,
  title={Ip-adapter: Text compatible image prompt adapter for text-to-image diffusion models},
  author={Ye, Hu and Zhang, Jun and Liu, Sibo and Han, Xiao and Yang, Wei},
  journal={arXiv preprint arXiv:2308.06721},
  year={2023}
}

@article{chen2023photoverse,
  title={Photoverse: Tuning-free image customization with text-to-image diffusion models},
  author={Chen, Li and Zhao, Mengyi and Liu, Yiheng and Ding, Mingxu and Song, Yangyang and Wang, Shizun and Wang, Xu and Yang, Hao and Liu, Jing and Du, Kang and others},
  journal={arXiv preprint arXiv:2309.05793},
  year={2023}
}

@article{xiao2025fastcomposer,
  title={Fastcomposer: Tuning-free multi-subject image generation with localized attention},
  author={Xiao, Guangxuan and Yin, Tianwei and Freeman, William T and Durand, Fr{\'e}do and Han, Song},
  journal={International Journal of Computer Vision},
  volume={133},
  number={3},
  pages={1175--1194},
  year={2025},
  publisher={Springer}
}

@article{zhang2024flashface,
  title={Flashface: Human image personalization with high-fidelity identity preservation},
  author={Zhang, Shilong and Huang, Lianghua and Chen, Xi and Zhang, Yifei and Wu, Zhi-Fan and Feng, Yutong and Wang, Wei and Shen, Yujun and Liu, Yu and Luo, Ping},
  journal={arXiv preprint arXiv:2403.17008},
  year={2024}
}

@article{huang2024context,
  title={In-context lora for diffusion transformers},
  author={Huang, Lianghua and Wang, Wei and Wu, Zhi-Fan and Shi, Yupeng and Dou, Huanzhang and Liang, Chen and Feng, Yutong and Liu, Yu and Zhou, Jingren},
  journal={arXiv preprint arXiv:2410.23775},
  year={2024}
}

@article{wang2024instantid,
  title={Instantid: Zero-shot identity-preserving generation in seconds},
  author={Wang, Qixun and Bai, Xu and Wang, Haofan and Qin, Zekui and Chen, Anthony and Li, Huaxia and Tang, Xu and Hu, Yao},
  journal={arXiv preprint arXiv:2401.07519},
  year={2024}
}

@article{xie2023omnicontrol,
  title={Omnicontrol: Control any joint at any time for human motion generation},
  author={Xie, Yiming and Jampani, Varun and Zhong, Lei and Sun, Deqing and Jiang, Huaizu},
  journal={arXiv preprint arXiv:2310.08580},
  year={2023}
}

@inproceedings{chen2025unireal,
  title={Unireal: Universal image generation and editing via learning real-world dynamics},
  author={Chen, Xi and Zhang, Zhifei and Zhang, He and Zhou, Yuqian and Kim, Soo Ye and Liu, Qing and Li, Yijun and Zhang, Jianming and Zhao, Nanxuan and Wang, Yilin and others},
  booktitle={Proceedings of the Computer Vision and Pattern Recognition Conference},
  pages={12501--12511},
  year={2025}
}

@article{cui2025emu3,
  title={Emu3. 5: Native Multimodal Models are World Learners},
  author={Cui, Yufeng and Chen, Honghao and Deng, Haoge and Huang, Xu and Li, Xinghang and Liu, Jirong and Liu, Yang and Luo, Zhuoyan and Wang, Jinsheng and Wang, Wenxuan and others},
  journal={arXiv preprint arXiv:2510.26583},
  year={2025}
}

@misc{flux1controlnet,
  author       = {InstantX Team},
  title        = {{Flux.1-dev-controlnet-union-pro}},
  howpublished = {\url{https://github.com/InstantX/Flux.1-dev-controlnet-union-pro}},
  note         = {Accessed: Nov. 2025},
  year         = {2025}
}

@article{ye2025echo,
  title={Echo-4o: Harnessing the power of gpt-4o synthetic images for improved image generation},
  author={Ye, Junyan and Jiang, Dongzhi and Wang, Zihao and Zhu, Leqi and Hu, Zhenghao and Huang, Zilong and He, Jun and Yan, Zhiyuan and Yu, Jinghua and Li, Hongsheng and others},
  journal={arXiv preprint arXiv:2508.09987},
  year={2025}
}

@article{xu2025withanyone,
  title={WithAnyone: Towards Controllable and ID Consistent Image Generation},
  author={Xu, Hengyuan and Cheng, Wei and Xing, Peng and Fang, Yixiao and Wu, Shuhan and Wang, Rui and Zeng, Xianfang and Jiang, Daxin and Yu, Gang and Ma, Xingjun and others},
  journal={arXiv preprint arXiv:2510.14975},
  year={2025}
}

@article{ku2023viescore,
  title={Viescore: Towards explainable metrics for conditional image synthesis evaluation},
  author={Ku, Max and Jiang, Dongfu and Wei, Cong and Yue, Xiang and Chen, Wenhu},
  journal={arXiv preprint arXiv:2312.14867},
  year={2023}
}

@article{xia2025dreamomni2,
  title={DreamOmni2: Multimodal Instruction-based Editing and Generation},
  author={Xia, Bin and Peng, Bohao and Zhang, Yuechen and Huang, Junjia and Liu, Jiyang and Li, Jingyao and Tan, Haoru and Wu, Sitong and Wang, Chengyao and Wang, Yitong and others},
  journal={arXiv preprint arXiv:2510.06679},
  year={2025}
}

@inproceedings{li2025openhumanvid,
  title={Openhumanvid: A large-scale high-quality dataset for enhancing human-centric video generation},
  author={Li, Hui and Xu, Mingwang and Zhan, Yun and Mu, Shan and Li, Jiaye and Cheng, Kaihui and Chen, Yuxuan and Chen, Tan and Ye, Mao and Wang, Jingdong and others},
  booktitle={Proceedings of the Computer Vision and Pattern Recognition Conference},
  pages={7752--7762},
  year={2025}
}

@article{nan2024openvid,
  title={Openvid-1m: A large-scale high-quality dataset for text-to-video generation},
  author={Nan, Kepan and Xie, Rui and Zhou, Penghao and Fan, Tiehan and Yang, Zhenheng and Chen, Zhijie and Li, Xiang and Yang, Jian and Tai, Ying},
  journal={arXiv preprint arXiv:2407.02371},
  year={2024}
}

@article{yuan2025opens2v,
  title={Opens2v-nexus: A detailed benchmark and million-scale dataset for subject-to-video generation},
  author={Yuan, Shenghai and He, Xianyi and Deng, Yufan and Ye, Yang and Huang, Jinfa and Lin, Bin and Luo, Jiebo and Yuan, Li},
  journal={arXiv preprint arXiv:2505.20292},
  year={2025}
}

@article{bai2025qwen2,
  title={Qwen2. 5-vl technical report},
  author={Bai, Shuai and Chen, Keqin and Liu, Xuejing and Wang, Jialin and Ge, Wenbin and Song, Sibo and Dang, Kai and Wang, Peng and Wang, Shijie and Tang, Jun and others},
  journal={arXiv preprint arXiv:2502.13923},
  year={2025}
}

@article{oquab2023dinov2,
  title={Dinov2: Learning robust visual features without supervision},
  author={Oquab, Maxime and Darcet, Timoth{\'e}e and Moutakanni, Th{\'e}o and Vo, Huy and Szafraniec, Marc and Khalidov, Vasil and Fernandez, Pierre and Haziza, Daniel and Massa, Francisco and El-Nouby, Alaaeldin and others},
  journal={arXiv preprint arXiv:2304.07193},
  year={2023}
}

@inproceedings{liu2024grounding,
  title={Grounding dino: Marrying dino with grounded pre-training for open-set object detection},
  author={Liu, Shilong and Zeng, Zhaoyang and Ren, Tianhe and Li, Feng and Zhang, Hao and Yang, Jie and Jiang, Qing and Li, Chunyuan and Yang, Jianwei and Su, Hang and others},
  booktitle={European conference on computer vision},
  pages={38--55},
  year={2024},
  organization={Springer}
}

@misc{flux1-fill-dev,
  author       = {Black Forest Labs},
  title        = {FLUX.1-Fill-dev},
  howpublished = {\url{https://huggingface.co/black-forest-labs/FLUX.1-Fill-dev}},
  note         = {Hugging Face model card},
  year         = {2025}
}

@misc{openai_gpt4_1_2025,
  author       = {OpenAI},
  title        = {GPT-4.1},
  year         = {2025},
  howpublished = {\url{https://openai.com/index/gpt-4-1}} 
}

@inproceedings{huang2025resolving,
  title={Resolving multi-condition confusion for finetuning-free personalized image generation},
  author={Huang, Qihan and Fu, Siming and Liu, Jinlong and Jiang, Hao and Yu, Yipeng and Song, Jie},
  booktitle={Proceedings of the AAAI Conference on Artificial Intelligence},
  volume={39},
  number={4},
  pages={3707--3714},
  year={2025}
}

@inproceedings{peebles2023scalable,
  title={Scalable diffusion models with transformers},
  author={Peebles, William and Xie, Saining},
  booktitle={Proceedings of the IEEE/CVF international conference on computer vision},
  pages={4195--4205},
  year={2023}
}

@article{cheng2025umo,
  title={UMO: Scaling Multi-Identity Consistency for Image Customization via Matching Reward},
  author={Cheng, Yufeng and Wu, Wenxu and Wu, Shaojin and Huang, Mengqi and Ding, Fei and He, Qian},
  journal={arXiv preprint arXiv:2509.06818},
  year={2025}
}

@article{ravi2024sam,
  title={Sam 2: Segment anything in images and videos},
  author={Ravi, Nikhila and Gabeur, Valentin and Hu, Yuan-Ting and Hu, Ronghang and Ryali, Chaitanya and Ma, Tengyu and Khedr, Haitham and R{\"a}dle, Roman and Rolland, Chloe and Gustafson, Laura and others},
  journal={arXiv preprint arXiv:2408.00714},
  year={2024}
}

@misc{schuhmann2024improvedaesthetic,
  author       = {Christoph Schuhmann},
  title        = {Improved Aesthetic Predictor},
  year         = {2024},
  howpublished = {\url{https://github.com/christophschuhmann/improved-aesthetic-predictor}},
  note         = {improved-aesthetic-predictor Lab},
  urldate      = {2025-11-13}
}
}

\section{Implementation Details about OpenSubject Data Construction Pipeline}
\subsection{Video Curation}
A critical step in constructing the \textbf{OpenSubject} dataset lies in the curation of high-quality video sources. To ensure both identity consistency and visual fidelity, we establish two key criteria: (1) the presence of a single, persistent subject throughout the video without scene transitions, and (2) high aesthetic quality with clear visual composition. We build upon the filtering pipelines adopted by several existing open-source video datasets, which have already performed coarse-level segmentation and initial quality screening. Based on this foundation, we select three high-quality video datasets, OpenHumanVid~\cite{li2025openhumanvid}, OpenVid~\cite{nan2024openvid}, and OpenS2V~\cite{yuan2025opens2v}, as our primary video sources. Details are as follows.

\noindent \textbf{OpenHumanVid} offers three notable advantages. (1) \emph{Large scale and high resolution:} it contains a wide variety of 720p–1080p videos collected from professional media sources such as films, television series, and documentaries, covering diverse human appearances, poses, expressions, and environments. (2) \emph{Rich motion conditions:} it includes multiple motion-driven modalities such as text prompts, skeletal trajectories, and synchronized speech audio, supporting both short and long textual forms. (3) \emph{Comprehensive identity representation:} the dataset encompasses individuals with varied demographics and appearance diversity, facilitating research on identity preservation and controllable generation.

\noindent \textbf{OpenVid-1M} complements OpenHumanVid by providing over one million high-quality text–video pairs, each accompanied by expressive and contextually rich captions. Its open-domain coverage makes it well-suited for training text-to-video (T2V) models under general scenarios.

\noindent \textbf{OpenS2V} contributes five million 720p subject–text–video triples, emphasizing fine-grained subject representation. Specifically, subject diversity is enhanced by (1) segmenting subjects and constructing cross-video associations to capture identity coherence, and (2) prompting GPT-Image on raw frames to synthesize multi-view depictions. However, OpenS2V relies on synthetic generation of subject reference images and restricts to single-subject cases, which motivates us to extend beyond its design to support multi-subject and manipulation-oriented tasks.

After collecting these datasets, we perform an additional quality screening step. Given that their intra-video subject consistency is already satisfactory, no further scene segmentation is required. We directly remove low-resolution (below 720P) or low-aesthetic (below 5.8) samples using the \textbf{aesthetic predictor}~\cite{schuhmann2024improvedaesthetic}, ensuring that only visually pleasing and identity-consistent clips are retained.

\subsection{Cross-Frame Subject Mining and Pairing}
To construct reliable subject–reference pairs from raw videos, we perform cross-frame subject mining followed by multi-stage verification to ensure identity consistency, completeness, and visual quality.

\paragraph{Frame Sampling.}
Processing all frames from each video would incur prohibitive computational cost and is unnecessary for discovering stable subject appearances. Instead, we adopt a \emph{sparse middle-frame sampling} strategy: for every video, we uniformly extract four frames from the central region of the sequence. This strategy provides a good balance between efficiency and reliability—middle frames typically avoid transitional or unstable segments, while four-frame sampling offers sufficient temporal diversity for cross-frame verification.

\paragraph{Cross-Frame Subject Verification.}
Given the sampled frames, we employ \textbf{Qwen2.5-VL-7B}~\cite{bai2025qwen2} to detect all visible foreground objects. For each video, we retain only those objects that appear consistently across at least two of the sampled frames. This cross-frame intersection step eliminates transient or low-confidence detections, ensuring that the remaining objects represent stable, trackable subjects suitable for subject-driven generation and manipulation tasks. The prompt is shown in Fig.~\ref{fig:prompt_object_extraction}.
\begin{figure*}[t!]
    \centering
    \begin{tcolorbox}[width=\textwidth]
        **Role**\\  
        You are a senior visual analyst. \\ 
        
        **Task**\\  
        List every main object that\\  
        1. covers $\geq 5 \%$ of the image area.\\  
        2. is $\geq 50 \%$ visible to the naked eye.  \\
        
        **Naming rule**\\  
        - Use single, generic, singular nouns in lower-case English, such as: “person”, “dog”, “car”, “chair”, “backpack”, …  \\
        - Skip brand, model, color, material, adjectives. \\ 
        - One distinct physical entity = one string.\\
        - Provide the names of up to three objects.  \\
        
        **Output format (strict JSON, no comments, no markdown wrappers)** \\ 
        \{``objects": [$``<name1>", ``<name2>"$, ...]\}
    \end{tcolorbox}
    \caption{Prompt for object extraction.}
    \label{fig:prompt_object_extraction}
\end{figure*}

\paragraph{Local Verification.}
While cross-frame overlap ensures temporal consistency, it does not guarantee subject quality. Therefore, we introduce a two-stage local verification process:

\begin{itemize}
    \item \textbf{Role-based filtering.}  
    We first remove subjects that do not meet pre-defined semantic roles (e.g., human, pet, common manipulable objects). Although effective for rejecting irrelevant categories, this step is insufficient for handling real-world visual defects such as occlusion, truncation, motion blur, or missing body parts.
    
    \item \textbf{VLM-based quality filtering.}  
    To address these cases, we further apply a Vision-Language Model (VLM) \textbf{Qwen2.5-VL-7B} evaluator to assess subject completeness and visibility. The VLM checks for occlusion level, degree of blur, boundary completeness, and viewpoint stability. Subjects failing these checks are discarded. This step substantially improves subject reliability compared to role-based filtering alone.
\end{itemize}

\subsubsection{Role-based filtering}
\label{sec:filter}

We perform two cleaning pipeline on the human-only and mixing videos to retain only high-quality training samples. 

\paragraph{(1) Human-only Clips.} The role-based filtering for human only are as follows.

\noindent \textbf{1) Single-Person Clips.}
For sequences that contain only one bounding box per frame, a sample is accepted if and only if it simultaneously satisfies
\begin{align}
N &= 1, \label{eq:single-n}\\
0.2 \leq A_{\text{box}} &\leq 0.8, \label{eq:single-area}\\
\operatorname{conf} &\geq 0.85, \label{eq:single-conf}\\
\bigl(\,\text{top}\in\mathcal{B} \land \text{bottom}\in\mathcal{B} \land A_{\text{box}}>0.7\,\bigr) &= \text{false}, \label{eq:single-through}
\end{align}
where $A_{\text{box}}=w\cdot h$ is the normalized area and $\mathcal{B}$ denotes the set of border tags (Sec.~\ref{sec:border}).

\noindent \textbf{2) 2-3 Person Clips}
For three-person scenes we require
\begin{align}
N &= 3, \label{eq:three-n}\\
\operatorname{IoU}(b_i,b_j) &\leq 0.2 \quad \forall\,0\leq i<j<3, \label{eq:three-iou}\\
\sum\nolimits_{k=0}^{2}A_{\text{box}}^{(k)} &\geq 0.2, \label{eq:three-sum}\\
\operatorname{conf}^{(k)} &\geq 0.8 \quad \forall\,k, \label{eq:three-conf}
\end{align}
together with the border--confidence rules summarised in Table~\ref{tab:border}.

\begin{table}[t]
    \centering
    \small
    \caption{Border--confidence constraints for three-person samples. $\mathcal{B}$ is the border set of a box. Any box that triggers the left condition must satisfy the minimum confidence on the right; otherwise the whole sample is rejected.}
    \label{tab:border}
    \begin{tabular}{@{}ll@{}}
        \toprule
        Triggering condition & Min.\ conf. \\ \midrule
        $\text{bottom}\in\mathcal{B}\land(\text{left}\in\mathcal{B}\lor\text{right}\in\mathcal{B})$ & 0.87 \\
        $\{\text{top},\text{bottom},\text{left}\}\subseteq\mathcal{B}$ \textbf{or} $\{\text{top},\text{bottom},\text{right}\}\subseteq\mathcal{B}$ & 0.88 \\
        $\{\text{top},\text{bottom}\}\subseteq\mathcal{B}$ & 0.87 \\
        $\text{top}\in\mathcal{B}\land\text{bottom}\notin\mathcal{B}$ & \emph{reject} \\ \bottomrule
    \end{tabular}
\end{table}

\noindent \textbf{3) Border Tagging}
\label{sec:border}
A box $(x_c,y_c,w,h)$ is converted to pixel coordinates as
\begin{equation}
\begin{aligned}
x_1 &= (x_c-\tfrac{w}{2})W, &\quad
y_1 &= (y_c-\tfrac{h}{2})H,\\
x_2 &= (x_c+\tfrac{w}{2})W, &\quad
y_2 &= (y_c+\tfrac{h}{2})H.
\end{aligned}
\end{equation}

With tolerance $\tau=15$\,px, the border set is defined as
\begin{equation}
\mathcal{B}=
\bigl\{\beta\in\{\text{top},\text{bottom},\text{left},\text{right}\}\mid d_\beta\leq\tau\bigr\},
\end{equation}
with $d_{\text{top}}=y_1$, $d_{\text{bottom}}=H-y_2$, $d_{\text{left}}=x_1$, and $d_{\text{right}}=W-x_2$.

\noindent \textbf{4) Filtering Algorithm}
The complete filtering procedure is summarised in Algorithm~\ref{alg:filter}.
After processing, \emph{kept} samples are re-written to the original \texttt{jsonl} files, while discarded records are permanently removed.

\begin{algorithm}[t]
    \small
    \caption{Sample filtering for human use.}
    \label{alg:filter}
    \begin{algorithmic}[1]
        \Require JSONL file list $\mathcal{F}$, tolerance $\tau=15$\,px
        \For{$f\in\mathcal{F}$}
            \State $\mathcal{L}\gets\emptyset$
            \For{line $\ell$ in $f$}
                \State parse $(N,\{b_k\},\{c_k\},W,H)$ from $\ell$
                \If{$W\neq 1280$ \textbf{ or } $H\neq 720$}
                    \State \textbf{continue}
                \EndIf
                \If{$N=1$ \textbf{ and } \textsc{SingleValid}($b_0,c_0$)}
                    \State $\mathcal{L}\gets\mathcal{L}\cup\{\ell\}$
                \ElsIf{$N=3$ \textbf{ and } \textsc{ThreeValid}($\{b_k\},\{c_k\}$)}
                    \State $\mathcal{L}\gets\mathcal{L}\cup\{\ell\}$
                \EndIf
            \EndFor
            \State overwrite $f$ with $\mathcal{L}$
        \EndFor
        \State
        \Function{SingleValid}{$b,c$}
            \State $A\gets w\cdot h$ from $b$
            \If{$A<0.2$ \textbf{ or } $A>0.8$ \textbf{ or } $c<0.85$}
                \State \Return \textbf{false}
            \EndIf
            \State $\mathcal{B}\gets$\textsc{Border}($b$)
            \If{$\{\text{top},\text{bottom}\}\subseteq\mathcal{B}$ \textbf{ and } $A>0.7$}
                \State \Return \textbf{false}
            \EndIf
            \State \Return \textbf{true}
        \EndFunction
        \State
        \Function{ThreeValid}{$\{b_k\},\{c_k\}$}
            \For{$i<j$}
                \If{$\operatorname{IoU}(b_i,b_j)>0.2$}
                    \State \Return \textbf{false}
                \EndIf
            \EndFor
            \State $A_{\text{sum}}\gets\sum_k w_k h_k$
            \If{$A_{\text{sum}}<0.2$}
                \State \Return \textbf{false}
            \EndIf
            \For{$k=0,\dots,2$}
                \If{$c_k<0.8$}
                    \State \Return \textbf{false}
                \EndIf
                \State $\mathcal{B}\gets$\textsc{Border}($b_k$)
                \If{$\text{top}\in\mathcal{B}$ \textbf{ and } $\text{bottom}\notin\mathcal{B}$}
                    \State \Return \textbf{false}
                \EndIf
                \State \textbf{if} Table~\ref{tab:border} conditions are violated \textbf{ then return false}
            \EndFor
            \State \Return \textbf{true}
        \EndFunction
    \end{algorithmic}
\end{algorithm}

\paragraph{(2) Mixed-Object Clips.}
We apply a role-based, per-frame filtering strategy to mixed-object clips.
A lightweight cleaning stage removes low-quality detections according to the rules below, and the operations are executed in the order summarised in Algorithm~\ref{alg:filter-mixed}.

\noindent \textbf{1) Hard Rules.}

\begin{enumerate}[leftmargin=*,noitemsep,topsep=0pt]
  \item \textbf{Area}: $0.01 \leq A_{\text{box}} \leq 0.60$.
  \item \textbf{Confidence}: $\operatorname{conf}\geq 0.5$ for generic classes; $\operatorname{conf}\geq 0.8$ for \texttt{person}.
  \item \textbf{Person count}: $1 \leq \#\texttt{person} \leq 3$; frames without any \texttt{person} are discarded.
  \item \textbf{Total objects}: $1 \leq \text{object\_number} \leq 5$.
  \item \textbf{Singleton box}: if only one object remains, its area must lie in $[0.20, 0.60]$.
\end{enumerate}

\noindent \textbf{2) Blacklisted Categories.}
We discard any box whose label belongs to the blacklist in Table~\ref{tab:blacklist}.
These categories correspond to background elements, clothing parts, or furniture that rarely constitute valid human-centric training targets.

\begin{table}[t]
\centering
\small
\caption{Blacklisted categories used for filtering mixed-object clips.}
\label{tab:blacklist}
\begin{tabular}{@{}llll@{}}
\toprule
Category & Category & Category & Category \\ \midrule
armchairs & apron      & beard      & bench      \\
blouse    & building   & cabinet    & ceiling    \\
chair     & chest      & cityscape  & coat       \\
collar    & counter    & countertop & couch      \\
desk      & face       & faucet     & field      \\
finger    & foot       & hair       & hand       \\
head      & jersey     & jacket     & jumpsuit   \\
leggings  & neck       & pants      & podium     \\
scarf     & shirt      & shorts     & sky        \\
suit      & sweater    & table      & tire       \\
trousers  & t-shirt    & uniform    & vest       \\
wheel     & wetsuit    &            &           \\
\bottomrule
\end{tabular}
\end{table}

\begin{figure*}[t]
\centering
\includegraphics[width=\linewidth]{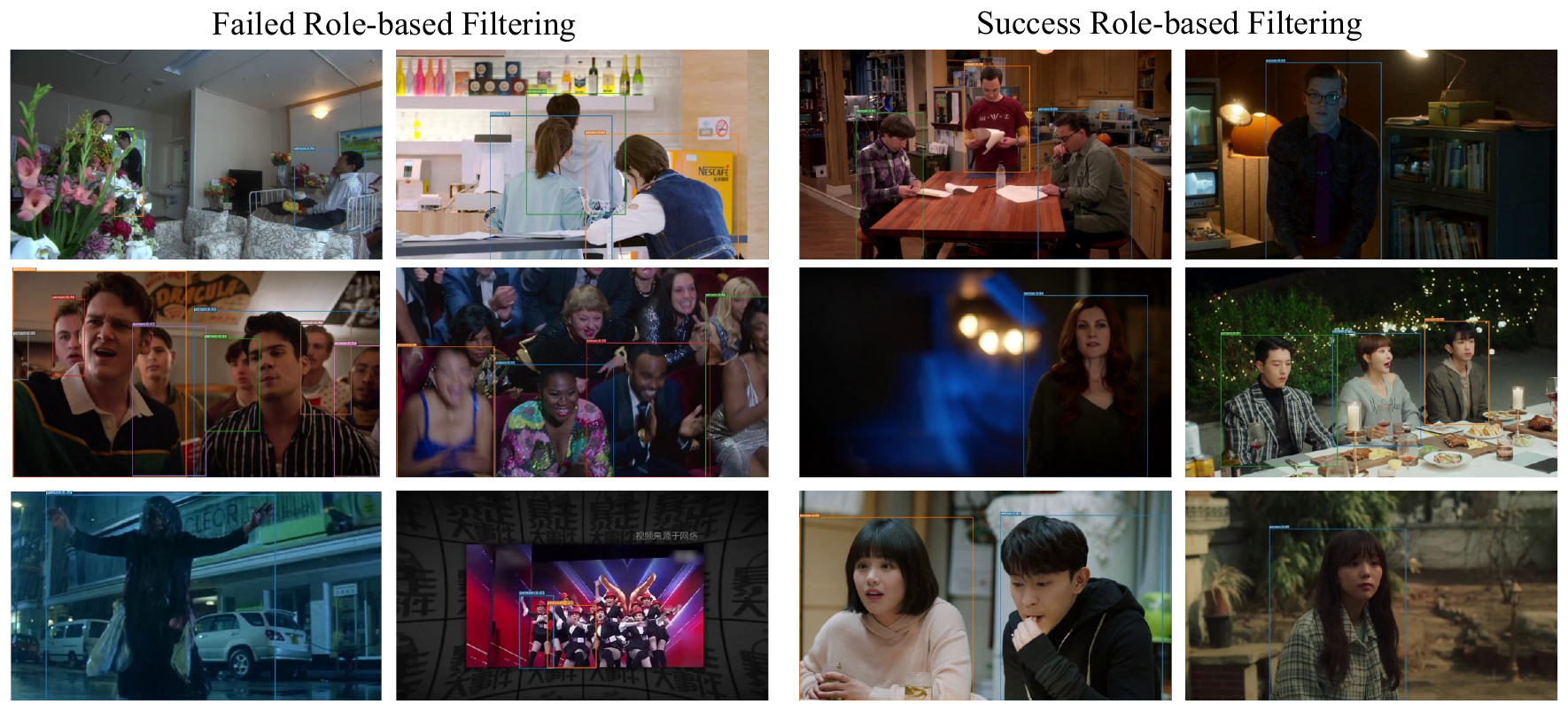}
\caption{Visualization of samples that fail and pass role-based bounding box filtering.}
\label{role_based_filtering}
\end{figure*}

\noindent \textbf{3) Duplicate Removal.}
For detections sharing the same label, we keep the one with the largest bounding-box area (line~\ref{line:dup} in Algorithm~\ref{alg:filter-mixed}). Some examples are shown in Fig.~\ref{role_based_filtering}.

\begin{algorithm}[t]
\small
\caption{Per-frame filtering for mixed-object clips.}
\label{alg:filter-mixed}
\begin{algorithmic}[1]
\Require raw labels $\mathcal{L}$, boxes $\mathcal{B}$, confidences $\mathcal{C}$
\Ensure cleaned sample or $\varnothing$
\State remove all boxes with $A\notin[0.01,0.60]$ \Comment{area filter}
\State remove all boxes whose label is in Table~\ref{tab:blacklist}
\State remove all boxes with $\text{conf}<0.5$
\For{each box $b$ with label \texttt{person}}
    \If{$\text{conf}(b)<0.8$}
        \State remove $b$
    \EndIf
\EndFor
\If{$\#\texttt{person}=0$ \textbf{ or } $\#\texttt{person}>3$}
    \State \Return $\varnothing$
\EndIf
\State merge duplicates by keeping the largest box per label \label{line:dup}
\State $N \gets |\mathcal{L}|$
\If{$N<1$ \textbf{ or } $N>5$}
    \State \Return $\varnothing$
\EndIf
\If{$N=1$}
    \State $A\gets\operatorname{area}(\mathcal{B}_0)$
    \If{$A<0.20$ \textbf{ or } $A>0.60$}
        \State \Return $\varnothing$
    \EndIf
\EndIf
\State \Return updated sample
\end{algorithmic}
\end{algorithm}

\subsubsection{VLM-based Filtering}
Beyond the rule-based constraints above, we further employ a VLM to filter out semantically unreliable detections.
Unlike the geometric and confidence thresholds, this stage explicitly verifies whether each cropped region visually matches its predicted label.

We treat human and non-human objects differently.
For boxes labelled as \texttt{person}, the VLM is prompted to describe the cropped region and to answer whether it depicts a complete, clearly visible human subject.
Only regions that are recognised as a person with sufficient detail (e.g., no severe occlusion, truncation, or motion blur) are kept; others are discarded.
For non-human objects, the VLM is asked to confirm both the presence and the category of the object and to reject regions that it recognizes as background textures or irrelevant accessories.

\begin{figure*}[t]
\centering
\includegraphics[width=\linewidth]{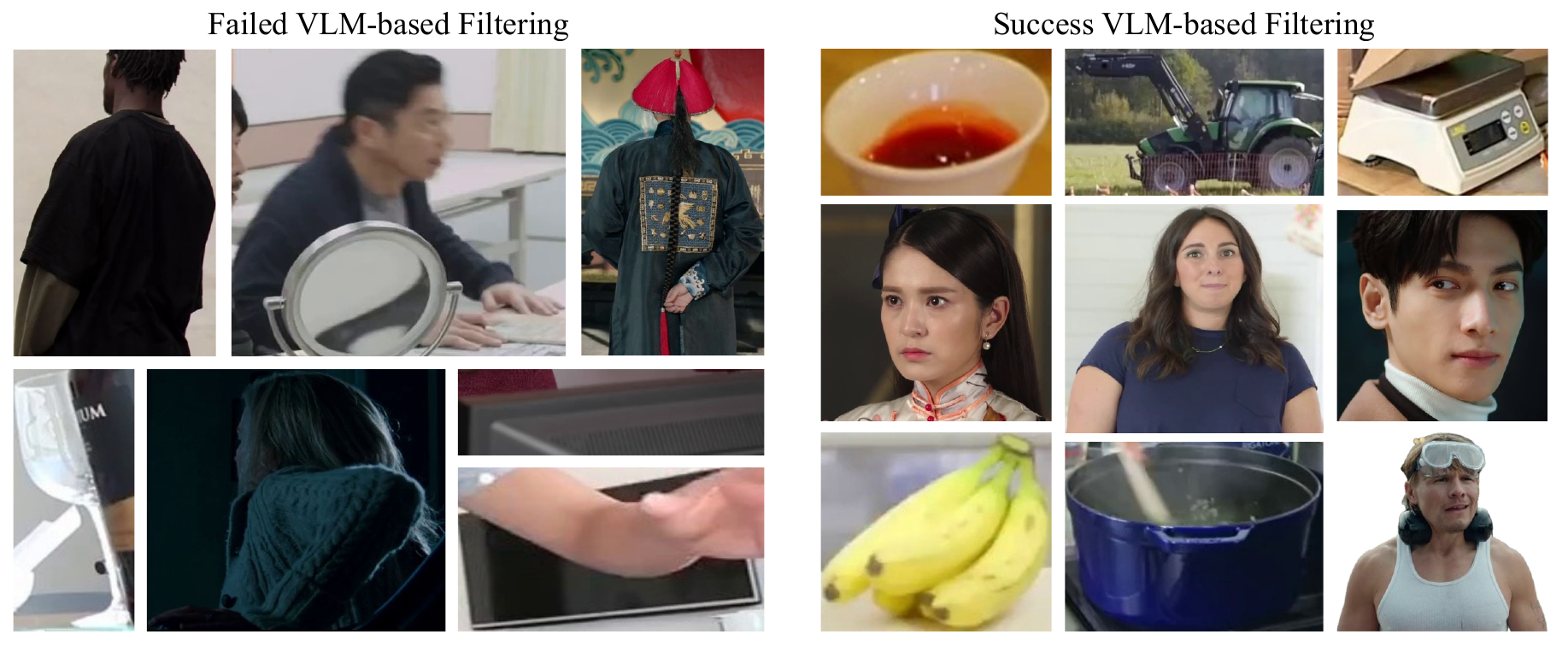}
\caption{Visualization of samples that fail and pass vlm-based bounding box filtering.}
\label{vlm_based_filtering}
\end{figure*}

The exact prompts used for human-centric and generic object filtering are illustrated in Fig.~\ref{fig:prompt_vlm_filtering_human} and Fig.~\ref{fig:prompt_vlm_filtering_object}, respectively.
These prompts are designed to be short and instruction-like, so that the VLM can robustly distinguish valid training targets from noisy detections in a wide range of scenes. Some examples are shown in Fig.~\ref{vlm_based_filtering}.

\begin{figure*}[t!]
    \centering
    \begin{tcolorbox}[width=\textwidth]
    *Task*:\\
    You are a quality-inspection agent. Given the cropped image of a single bounding-box labeled “person”, evaluate it against the following four criteria and return the exact JSON below.\\
    
    *Criteria*:\\
    Occlusion - if the body/face is heavily occluded ($\geq 15 \%$ hidden) → true, otherwise false \\
    Back - if the person is a back view → true, otherwise false \\
    Motion-blur - if the person region shows noticeable motion blur or camera shake $→$ true, otherwise false \\
    Full-face - if the full head is in frame → true, otherwise false \\
    Single-person - if the crop contains exactly one individual → true, otherwise false \\
    
    *Output Format: do not output any other text except the JSON format*:\\
    \{``Occlusion": true/false, ``Back": true/false, ``Motion-blur": true/false, ``Full-face": true/false, ``Single-person": true/false\}
    \end{tcolorbox}
    \caption{Prompt for VLM-based filtering for human.}
    \label{fig:prompt_vlm_filtering_human}
\end{figure*}
\begin{figure*}[t!]
    \centering
    \begin{tcolorbox}[width=\textwidth]
    *Task*:\\
    You are a quality-inspection agent. Given the cropped image of a single bounding-box labeled “person”, evaluate it against the following four criteria and return the exact JSON below.\\
    
    *Criteria*:\\
    Category-match - the main object must match the given label exactly.
    Completeness - the object must not be cut off by the image border ($\leq 10 \%$ occlusion allowed).
    Clarity - the object region must be free from motion blur, out-of-focus, heavy pixelation, under-exposure or over-exposure that hides recognizable details.
    Occlusion - if the body/face is heavily occluded ($\geq 15 \%$ hidden) → true, otherwise false \\
    Motion-blur - if the person region shows noticeable motion blur or camera shake $→$ true, otherwise false \\
    
    *Output Format: do not output any other text except the JSON format*:\\
    \{``category-match": true/false, ``completeness": true/false, ``clarity": true/false, ``Occlusion": true/false,``Motion-blur": true/false\}
    \end{tcolorbox}
    \caption{Prompt for VLM-based filtering for object.}
    \label{fig:prompt_vlm_filtering_object}
\end{figure*}

\begin{figure*}[t]
\centering
\includegraphics[width=\linewidth]{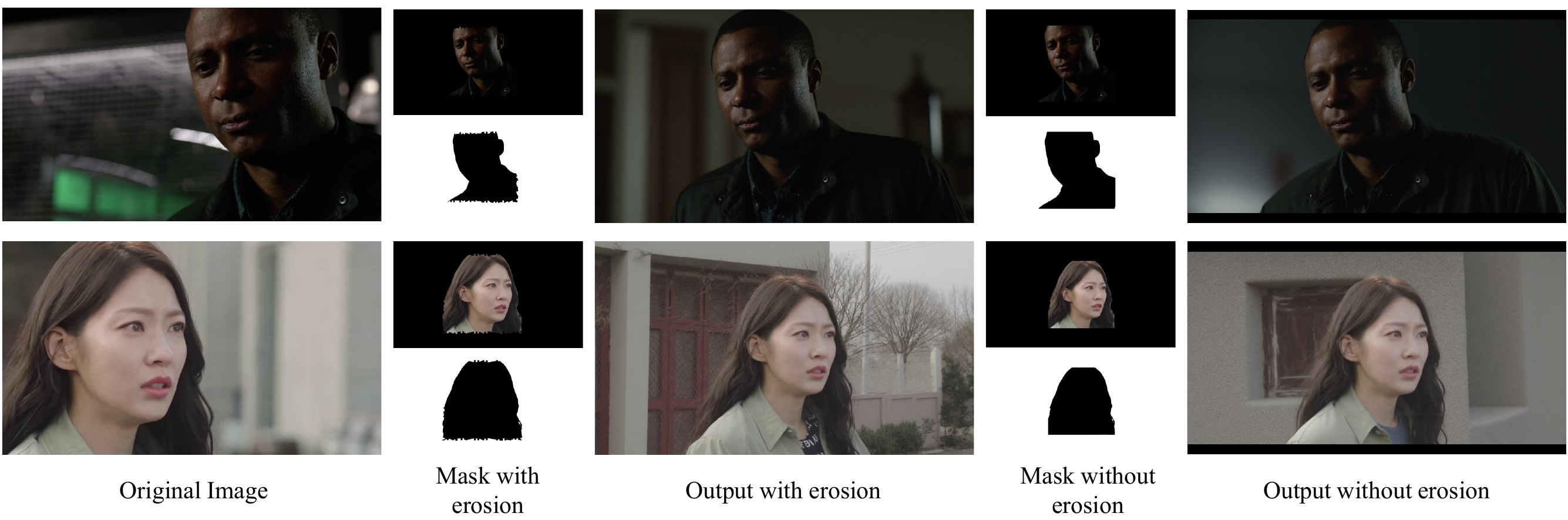}
\caption{Visualization of samples with and without erosion.}
\label{rescale}
\end{figure*}

\begin{figure*}[t]
\centering
\includegraphics[width=\linewidth]{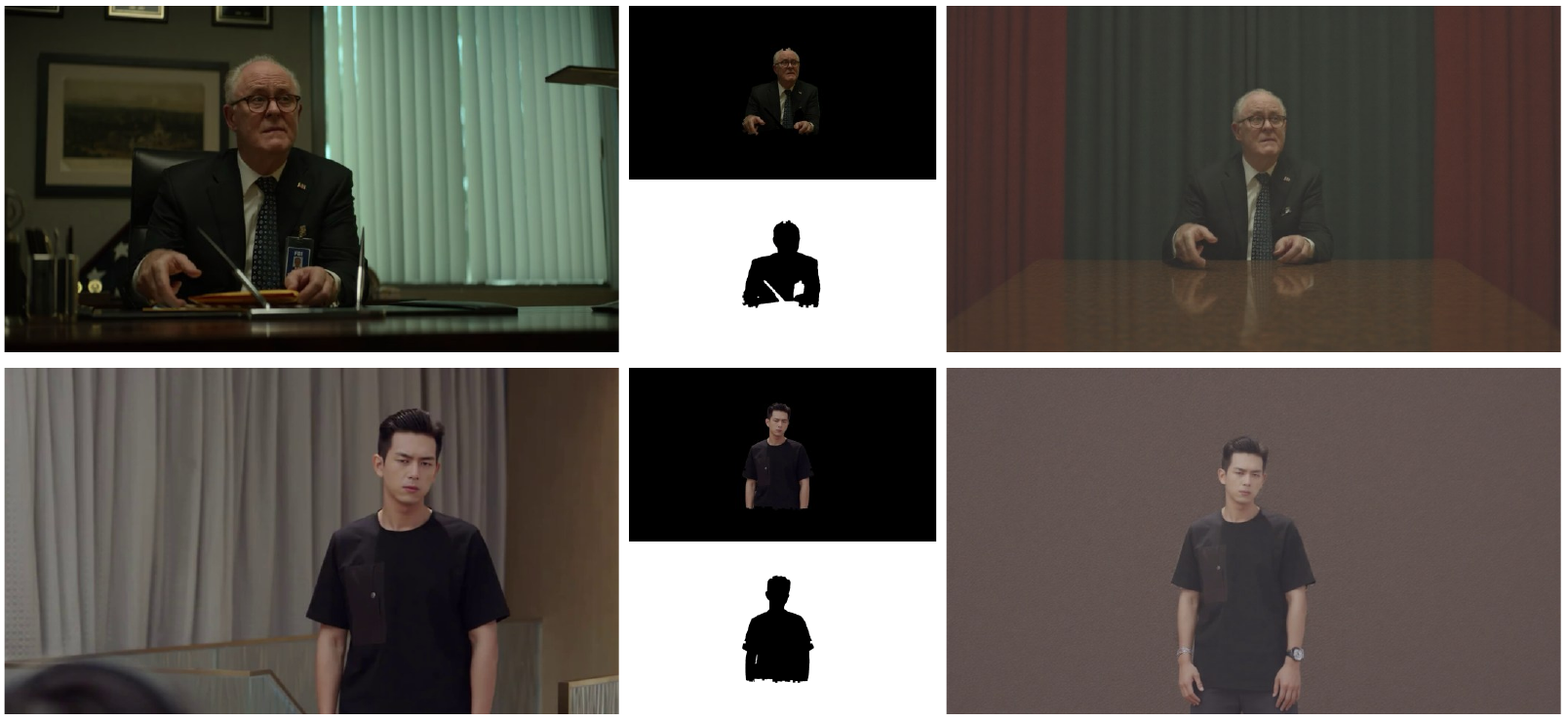}
\caption{Visualization of samples synthesized with FLUX.1-Fill [Dev].}
\label{pipeline_case}
\end{figure*}

\subsection{Identity-Preserving Reference Image Synthesis}
To construct reference images for subject-driven generation, we synthesize identity-preserving inputs via outpainting.
A naive choice is to use the detected bounding box as the outpainting region directly; however, object shapes are highly irregular, and the rectangular box typically includes large portions of irrelevant background.
For human subjects in particular, the background inside the bounding box provides a strong prior for the outpainting model~\cite{flux1-fill-dev}, which tends to reconstruct a scene that closely matches the original context, yielding reference images whose background is almost identical to the input frame.

To mitigate this issue, we instead derive an outpainting mask from the segmentation map and use it as a more precise structural guide.
The segmentation mask focuses the synthesis on the true foreground region, reducing background leakage.
For convenience and consistency with other components, we still associate each segmentation mask with its corresponding bounding box and optionally truncate the mask within that box.
In some cases (e.g., half-body crops), this truncation interacts with the prior of FLUX.1-Fill [Dev]~\cite{flux1-fill-dev}, producing unnaturally clean horizontal boundaries that manifest as black bars near the top or bottom of the synthesized image.
We therefore apply an irregular morphological erosion to the mask boundaries, introducing slight shape perturbations that break these artificial edges and lead to more natural, identity-preserving reference images.

Given an annotated frame $(I, \mathcal{M})$ with $N$ person instances, we generate an \emph{outpainting} pair $(\tilde{I}, \tilde{M})$ where the human figure is randomly re-positioned inside a $1280\!\times\!720$ canvas and the remaining region becomes the inpainting mask.
The pipeline is summarised in Algorithm~\ref{alg:outpaint} and detailed below.

\noindent \textbf{Notation.}
Let $b_i\!=\![x_c,y_c,w,h]$ be the \emph{normalised} box of instance $i$, and $M_i\!\in\!\{0,1\}^{H\times W}$ its binary mask.
The pixel-level foreground area is $A_i\!=\!\sum_{u,v}M_i[u,v]$.
The \emph{total} image area is $A_{\text{img}}\!=\!HW$.

\noindent \textbf{Step-1: overlap cleaning.}
To avoid nested detections we sequentially remove pixels of \emph{smaller} instances from \emph{larger} ones:
\begin{equation}
M_i'\!=\!M_i\,\setminus\!\bigcup_{j\,:\,A_j<A_i}\!M_j.
\end{equation}
After cleaning, each crop contains \emph{only} pixels belonging to the current person.

\noindent \textbf{Step-2: scale decision.}
An instance is regarded as \emph{small} if $A_i/A_{\text{img}}\!<\!0.30$.
For small targets, we \emph{up-scale} them so that their foreground occupies $30\%$--$40\%$ of the canvas; otherwise, we \emph{down-scale} by a random factor in $[0.6,0.8]$.
The scaling factor is therefore
\begin{equation}
s_i\!=\!
\begin{cases}
\sqrt{t A_{\text{img}}/A_i}, & \text{small object},\;t\!\sim\!\mathcal{U}(0.30,0.40),\\[4pt]
\mathcal{U}(0.6,0.8), & \text{otherwise}.
\end{cases}
\end{equation}
The scaled crop must fit the canvas, \emph{i.e.} $s_i w_i\!\leq\!W,\,s_i h_i\!\leq\!H$.

\noindent \textbf{Step-3: placement}
The top-left corner $(x_0,y_0)$ is sampled from a Gaussian centred at the image middle with $\sigma\!=\!0.1\,W$ (or $0.1\,H$), ensuring the object rarely touches the boundary.

\noindent \textbf{Step-4: mask generation}
The \emph{hole} mask $\tilde{M}\!\in\!\{0,255\}^{H\times W}$ is defined as
\begin{equation}
\tilde{M}[u,v]\!=\!
\begin{cases}
0, & \text{if }(u,v)\text{ lies inside the placed foreground},\\[2pt]
255, & \text{otherwise}.
\end{cases}
\end{equation}
Optionally, a stochastic \emph{tear-border} erosion (depth $\delta\!\sim\![5,25]$ px, frequency 15 px) can be applied to mimic hand-drawn rough edges. Some examples are shown in Fig.~\ref{rescale}.

\begin{algorithm}[t]
\small
\caption{Per-frame out-painting pair generation.}
\label{alg:outpaint}
\begin{algorithmic}[1]
\Require{image $I$, masks $\{M_i\}$, boxes $\{b_i\}$, canvas size $(W,H)$}
\Ensure{out-painted image $\tilde{I}$, hole mask $\tilde{M}$}
\State compute areas $A_i\!=\!\sum M_i$; sort descending
\For{$i=1\dots N$}
    \State $M_i'\!\gets\!M_i\setminus\bigcup_{j<i}M_j'$ \Comment{remove smaller overlaps}
    \State crop $C_i\!\gets\!I[b_i]$ masked by $M_i'$
    \If{$A_i/A_{\text{img}}<0.30$}
        \State $s\!\gets\!\sqrt{\mathcal{U}(0.30,0.40)\,A_{\text{img}}/A_i}$
    \Else
        \State $s\!\gets\!\mathcal{U}(0.6,0.8)$
    \EndIf
    \State $s\!\gets\!\min(s,\,W/w_i,\,H/h_i)$ \Comment{fit canvas}
    \State resize $(C_i,M_i')$ by $s$
    \State sample $(x_0,y_0)$ with Gaussian centre prior
    \State paste $C_i$ into $\tilde{I}$ at $(x_0,y_0)$
    \State set $\tilde{M}[x_0:x_0\!+\!s w_i,\,y_0:y_0\!+\!s h_i]\!=\!0$
\EndFor
\State \textbf{optional} apply tear-border erosion to $\tilde{M}$
\State \Return $(\tilde{I},\tilde{M})$
\end{algorithmic}
\end{algorithm}

The entire pipeline is implemented with OpenCV and NumPy and runs in parallel with a thread-pool (default 15 workers), achieving $\sim$650 fps on a single 64-core node. Some examples are shown in Fig.~\ref{pipeline_case}.

\subsection{Verification and Captioning}

To ensure that the extracted subjects and synthesized samples meet the quality requirements for subject-driven generation and manipulation, we employ Qwen2.5-VL-7B for both verification and caption generation. The model is prompted with carefully designed, task-specific instructions that assess visual quality, detect artifacts, and produce semantically rich descriptions.

For verification, we use structured prompts that guide the VLM to evaluate each cropped subject or synthesized sample across multiple dimensions, including completeness, visibility, geometric plausibility, artifact detection, and scene consistency. These prompts enable consistent, interpretable filtering decisions. Representative verification prompts are shown in Fig.~\ref{fig:prompt_vlm_artifact_plausibility}.

For captioning, we design separate prompt templates tailored for manipulation-focused and generation-focused tasks. Each template is available in both short-form and long-form variants to match different training needs. The short captions emphasize identity, appearance, and high-level context, while the long captions enrich the description with detailed attributes such as pose, viewpoint, clothing, background elements, and interaction cues. The captioning templates are illustrated in Fig.~\ref{fig:prompt_short_manipulation}, Fig.~\ref{fig:prompt_long_manipulation}, Fig.~\ref{fig:prompt_short_generation}, and Fig.~\ref{fig:prompt_long_generation}.

Together, these verification and captioning prompts form a comprehensive VLM-based pipeline that ensures high-quality subject selection and accurate textual descriptions, ultimately strengthening the reliability and diversity of the OpenSubject dataset.

\begin{figure*}[t!]
    \centering
    \begin{tcolorbox}[width=\textwidth]
    \text{*Task*:}\\
    You are an artifact-inspection agent. Given a synthesized image, evaluate whether the sample exhibits visual artifacts or violates basic physical plausibility. Follow the criteria below and return the required JSON only.\\
    
    \text{*Criteria*:}\\
    Geometry-error — true if there are distorted limbs, inconsistent proportions, broken body parts, or unnatural joint angles; otherwise false.\\
    Texture-artifact — true if there are repeated patterns, abnormal textures, over-smoothing, or patch-like regions; otherwise false.\\
    Lighting-violation — true if lighting direction, shadows, or reflections contradict physical plausibility; otherwise false.\\
    Background-conflict — true if the subject and background do not align (e.g., mismatched perspective, depth inconsistency, floating objects); otherwise false.\\
    
    \text{*Output Format: return only the JSON object below (no extra text)*}:\\
    \{
    ``Geometry-error": true/false,\\
    ``Texture-artifact": true/false,\\
    ``Lighting-violation": true/false,\\
    ``Background-conflict": true/false
    \}
    \end{tcolorbox}
    \caption{Prompt for VLM-based artifact and physical-plausibility assessment.}
    \label{fig:prompt_vlm_artifact_plausibility}
\end{figure*}

\begin{figure*}[t!]
    \centering
    \begin{tcolorbox}[width=\textwidth]
    \textbf{Prompt for short caption with generation style (manipulation task)} \\
    You are a professional, referenced subject replacement prompt writer.\\
    I will provide you with two input images and a output image: \\
    - Input Images: 1) Full scene photo (everything visible); 2) subject photo that must disappear \\
    - Output image: Identical scene, only the subject has been substituted.\\
    
    Your Task:\\
    1. Identify the difference between the complete scene image and the output image. The difference should be only the subject.\\
    2. In 1-2 sentences, describe how to swap the subject in the complete scene image to the output image. Pinpoint the swap process in the description. \\
    3. Since the subject photo is a part of the complete scene image, the complete scene image is not provided, you need to describe the details of the subject. \\
    4. Tone should mimic user's instruction prompt when using referenced subject replacement image generation model.\\
    5. Output only the description.\\

    \textbf{Prompt for short caption with editing style (manipulation task)} \\
    You are a professional, referenced subject replacement prompt writer. \\
    I will provide you with two input images and a output image: \\
    - Input Images: 1) Full scene photo (everything visible); 2) subject photo that must disappear \\
    - Output image: Identical scene, only the subject has been substituted. \\
    
    Your Task: \\
    1. Identify the difference between the complete scene image and the output image. The difference should be only the subject. \\
    2. In 1-2 sentences, describe how to edit the complete scene image to obtain the output image by substituting the subject. \\
    3. Since the subject photo is a part of the complete scene image, the complete scene image is not provided, you need to describe the details of the subject. \\
    4. Tone should mimic user's instruction prompt when using referenced subject replacement image editing model. \\
    5. Output only the description. \\
    \end{tcolorbox}
    \caption{Prompt for short caption with generation or editing style (manipulation task).}
    \label{fig:prompt_short_manipulation}
\end{figure*}
\begin{figure*}[t!]
    \centering
    \begin{tcolorbox}[width=\textwidth]
    \textbf{Prompt for long caption with generation style (manipulation task)} \\
    You are a professional, referenced subject replacement prompt writer.\\
    I will provide you with two input images and a output image: \\
    - Input Images: 1) Full scene photo (everything visible); 2) subject photo that must disappear \\
    - Output image: Identical scene, only the subject has been substituted.\\
    
    Your Task:\\
    1. Identify the difference between the complete scene image and the output image. The difference should be only the subject.\\
    2. In 3-4 sentences, describe how to swap the subject in the complete scene image to the output image. Pinpoint the swap process in the description. \\
    3. Since the subject photo is a part of the complete scene image, the complete scene image is not provided, you need to describe the details of the subject. \\
    4. Tone should mimic user's instruction prompt when using referenced subject replacement image generation model.\\
    5. Output only the description.\\

    \textbf{Prompt for long caption with editing style (manipulation task)} \\
    You are a professional, referenced subject replacement prompt writer. \\
    I will provide you with two input images and a output image: \\
    - Input Images: 1) Full scene photo (everything visible); 2) subject photo that must disappear \\
    - Output image: Identical scene, only the subject has been substituted. \\
    
    Your Task: \\
    1. Identify the difference between the complete scene image and the output image. The difference should be only the subject. \\
    2. In 3-4 sentences, describe how to edit the complete scene image to obtain the output image by substituting the subject. \\
    3. Since the subject photo is a part of the complete scene image, the complete scene image is not provided, you need to describe the details of the subject. \\
    4. Tone should mimic user's instruction prompt when using referenced subject replacement image editing model. \\
    5. Output only the description. \\
    \end{tcolorbox}
    \caption{Prompt for long caption with generation or editing style (manipulation task).}
    \label{fig:prompt_long_manipulation}
\end{figure*}
\begin{figure*}[t!]
    \centering
    \begin{tcolorbox}[width=\textwidth]
    \textbf{Prompt for short caption with generation style} \\
    You are a professional, multi-subject combination-driven new scene generation prompt writer. \\
    I will provide you with multiple images and tell you which one contains the target object: \\
    - Input Images: 1) Multiple images; 2)Label the objects (target objects) in the images. \\
    - Output image: A photograph of the combination of multiple objects (target objects) transplanted into a completely new scene. \\
    
    Your Task:\\
    1. Identify each object in the input images and its label, know that the objects in the input image is a part of the output image. \\
    2. In 1-2 sentences, describe how the objects in the input images are combined in the output image. Focus on the combination of multiple objects in the scene. \\
    3. Since the object images are cropped from the complete scene images,the complete scene image is not provided, you need to describe the details of the objects in the output image based on the information in the input images. Each object should be described in detail, such as the category/pose/appearance/color/material/dress-style/etc.\\
    4. Tone should mimic user's instruction prompt when using multi-subject combination-driven new scene generation image generation model. Should clearly describe the transformation of the object from input to output image.\\
    5. Output only the description.\\

    \textbf{Prompt for short caption with editing style} \\
    You are a professional, multi-subject combination-driven new scene generation prompt writer.\\
    I will provide you with multiple images and tell you which one contains the target object:\\
    - Input Images: 1) Multiple images; 2)Label the objects (target objects) in the images.\\
    - Output image: A photograph of the combination of multiple objects (target objects) transplanted into a completely new scene. \\
    
    Your Task:\\
    1. Identify each object in the input images and its label, know that the objects in the input image is a part of the output image.\\
    2. In 1-2 sentences, describe how the objects in the input images are combined in the output image. Focus on the combination of multiple objects in the scene.\\ 
    3. Since the object images are cropped from the complete scene images,the complete scene image is not provided, you need to describe the details of the objects in the output image based on the information in the input images. Each object should be described in detail, such as the category/pose/appearance/color/material/dress-style/etc.\\
    4. Tone should mimic user's instruction prompt when using multi-subject combination-driven new scene editing model. Should clearly describe the transformation of the object from input to output image.\\
    5. Output only the description.\\
    \end{tcolorbox}
    \caption{Prompt for short caption with generation or editing style.}
    \label{fig:prompt_short_generation}
\end{figure*}
\begin{figure*}[t!]
    \centering
    \begin{tcolorbox}[width=\textwidth]
    \textbf{Prompt for long caption with generation style} \\
    You are a professional, multi-subject combination-driven new scene generation prompt writer. \\
    I will provide you with multiple images and tell you which one contains the target object: \\
    - Input Images: 1) Multiple images; 2)Label the objects (target objects) in the images. \\
    - Output image: A photograph of the combination of multiple objects (target objects) transplanted into a completely new scene. \\
    
    Your Task:\\
    1. Identify each object in the input images and its label, know that the objects in the input image is a part of the output image. \\
    2. In 5-6 sentences, describe how the objects in the input images are combined in the output image. Focus on the combination of multiple objects in the scene. \\
    3. Since the object images are cropped from the complete scene images,the complete scene image is not provided, you need to describe the details of the objects in the output image based on the information in the input images. Each object should be described in detail, such as the category/pose/appearance/color/material/dress-style/etc.\\
    4. Tone should mimic user's instruction prompt when using multi-subject combination-driven new scene generation image generation model. Should clearly describe the transformation of the object from input to output image.\\
    5. Output only the description.\\

    \textbf{Prompt for long caption with editing style} \\
    You are a professional, multi-subject combination-driven new scene generation prompt writer.\\
    I will provide you with multiple images and tell you which one contains the target object:\\
    - Input Images: 1) Multiple images; 2)Label the objects (target objects) in the images.\\
    - Output image: A photograph of the combination of multiple objects (target objects) transplanted into a completely new scene. \\
    
    Your Task:\\
    1. Identify each object in the input images and its label, know that the objects in the input image is a part of the output image.\\
    2. In 5-6 sentences, describe how the objects in the input images are combined in the output image. Focus on the combination of multiple objects in the scene.\\ 
    3. Since the object images are cropped from the complete scene images,the complete scene image is not provided, you need to describe the details of the objects in the output image based on the information in the input images. Each object should be described in detail, such as the category/pose/appearance/color/material/dress-style/etc.\\
    4. Tone should mimic user's instruction prompt when using multi-subject combination-driven new scene editing model. Should clearly describe the transformation of the object from input to output image.\\
    5. Output only the description.\\
    \end{tcolorbox}
    \caption{Prompt for long caption with generation or editing style.}
    \label{fig:prompt_long_generation}
\end{figure*}

\section{Implementation Details about OpenSubject Benchmark}

In the main paper, we introduced the evaluation dimensions for both generation and manipulation tasks. Here, we provide additional details on the scoring procedure, rubric design, and prompt construction used in our benchmark.

\paragraph{Evaluation Protocol.}
Following recent instruction-based evaluation frameworks such as VIEScore~\cite{ku2023viescore} and OmniContext~\cite{wu2025omnigen2}, we employ a strong VLM judge---GPT-4.1~\cite{openai_gpt4_1_2025}---to assign 0--10 scores according to rubricized prompts with clearly defined criteria. Each dimension is evaluated independently to ensure controllable, fine-grained assessment.

\paragraph{Generation Tasks.}
For \textbf{generation} scenarios, we evaluate a model's ability to satisfy textual instructions and identity constraints. We report:
\begin{itemize}
    \item \textbf{Prompt Adherence (PA):} Measures how well the generated output satisfies textual attributes, object counts, and relational descriptions.
    \item \textbf{Identity Fidelity (IF):} Measures consistency of the generated subject with the provided reference images, including facial features, hairstyle, clothing cues, and global appearance.
    \item \textbf{Overall:} Defined as the geometric mean of PA and IF, reflecting a balanced evaluation of semantic accuracy and identity preservation.
\end{itemize}
The complete rubric and PA/IF scoring prompts are shown in Fig.~\ref{fig:prompt_pa_score} and Fig.~\ref{fig:prompt_if_score}.

\paragraph{Manipulation Tasks.}
For \textbf{manipulation} scenarios, we measure whether the model faithfully edits the target region without altering irrelevant content. We report:
\begin{itemize}
    \item \textbf{Manipulation Fidelity (MF):} Evaluates how accurately the edited region matches the requested change and, when applicable, the reference subject(s). This includes the correctness of replaced identity, edited attributes, and local semantics.
    \item \textbf{Background Consistency (BC):} Assesses whether regions outside the edited area remain unchanged, including background structure, lighting, object layout, and scene components.
    \item \textbf{Overall:} The geometric mean of MF and BC, reflecting the trade-off between edit accuracy and scene stability.
\end{itemize}
The MF and BC scoring prompts are presented in Fig.~\ref{fig:prompt_mf_score} and Fig.~\ref{fig:prompt_bc_score}.

\begin{figure*}[t!]
    \centering
    \begin{tcolorbox}[width=\textwidth]
    --- Setting \\
    You are given multiple images:\\
    - The **first image** is the **ground truth (reference)**, serving as a visual standard for assessing scene similarity.  \\
    - The **second image** is the **model's output image**, which you need to evaluate. \\
    
    --- Task: Rate from 0 to 10 \\
    Evaluate whether the **subject from the reference image** has been correctly and faithfully **manipulated or integrated** into the **base image**, in accordance with the **editing instruction**. \\
    
    ---- **Scoring Criteria** \\
    
    - **0:** The manipulation *completely failed* — the reference subject is missing or unrecognizable.  \\
    - **1-3:** The manipulation is *severely incorrect*, showing only vague or partial resemblance to the reference subject.  \\
    - **4-6:** The manipulation is *partially successful* — some recognizable traits of the reference subject appear, but major inconsistencies remain.  \\
    - **7-9:** The manipulation is *mostly accurate*, with strong resemblance to the reference subject and only minor mismatches.  \\
    - **10:** The manipulation is *perfectly successful* — the subject is seamlessly and accurately represented, matching the reference in identity, structure, and style. \\
    
    ---- **Pay special attention to** \\
    
    - **Identity fidelity:** Facial structure, hairstyle, clothing, and other distinctive features should closely match the reference subject.  \\
    - **Pose and spatial alignment:** The manipulated subject should align naturally with the scene's geometry, position, and orientation in the base image.  \\
    - **Expression and attributes:** Facial expressions and physical traits (e.g., age, gender, skin tone) should remain consistent with the reference. \\ 
    - **Semantic correctness:** The correct individual or object from the reference should appear exactly where specified by the instruction. \\ 
    - **Selective manipulation:** Only the target subject should be replaced or modified — no unintended entities should appear or disappear. \\
    
    ---- **Important Notes** \\
    
    - **Deduct points** for every visible mismatch in identity, shape, or manipulation accuracy.  \\
    - Do *not* consider **background consistency**, **artifact realism**, or **aesthetic appeal** — focus exclusively on the correctness of the subject manipulation itself. \\ 
    - The final score should reflect **how accurately and faithfully the model followed the instruction to manipulate or replace the subject**.  \\
    - **Scoring should be strict** — assign high scores only when the manipulated subject strongly and consistently matches the reference. \\
    
    **Editing instruction:** $`<instruction>`$\\
    \end{tcolorbox}
    \caption{Prompt for PA score.}
    \label{fig:prompt_pa_score}
\end{figure*}
\begin{figure*}[t!]
    \centering
    \begin{tcolorbox}[width=\textwidth]
    --- Setting \\
    You are given multiple images: \\
    - The **first image** is the **ground truth (reference)**, serving as a visual standard for assessing scene similarity.  \\
    - The **second image** is the **model's output image**, which you need to evaluate. \\
    
    --- Task: Rate from 0 to 10 \\
    Evaluate whether the **identities of subjects** in the **final image** match those of the corresponding individuals in the **first images**. \\
    
    ---- **Scoring Criteria** \\
    
    - **0:** The subject identities in the image are *completely inconsistent* with those in the reference images.  \\
    - **1-3:** The identities are *severely inconsistent*, showing only a few minor resemblances.  \\
    - **4-6:** The image displays *some notable similarities*, but major inconsistencies remain — indicating a *moderate* level of identity match.  \\
    - **7-9:** The identities are *mostly consistent*, with only subtle or localized mismatches.  \\
    - **10:** The subject identities in the final image are *perfectly consistent* with those in the reference images. \\
    
    ---- **Pay special attention to** \\
    
    - **Facial and cranial features:** Match in the appearance and placement of eyes, nose, mouth, cheekbones, jawline, wrinkles, makeup, hairstyle, hair color, and overall head shape.  \\
    - **Correct identity usage:** Verify that the proper individuals or objects from the input images are used (no identity swaps or omissions).  \\
    - **Physical traits:** Check that body shape, skin tone, and other defining physical attributes remain consistent, without distortion or abnormal anatomy.  \\
    - **Clothing and accessories:** If the instruction does *not* request changes to clothing or hairstyle, ensure these remain consistent with the input images.  \\
    - **Subtle identity cues:** Look for alignment in facial expression, proportions, and unique personal features (e.g., freckles, scars, glasses). \\
    
    ---- **Important Notes** \\
    
    - **Deduct points** for each visible identity mismatch or inconsistency. \\
    - **Deduct points** for each unreasonable lighting on the face. Please use the **first image** as a reference. \\
    - The score must reflect the **identity consistency** across all objects mentioned in the instruction, please use the first image as a reference.  \\
    - **Scoring should be strict** — high scores should only be given when identity match is clearly strong and consistent throughout. \\
    
    **Editing instruction:** $`<instruction>`$ \\
    \end{tcolorbox}
    \caption{Prompt for IF score.}
    \label{fig:prompt_if_score}
\end{figure*}
\begin{figure*}[t!]
    \centering
    \begin{tcolorbox}[width=\textwidth]
    --- Setting \\
    You are given multiple images: \\
    - The **first image** is the **ground truth (reference)**, serving as a visual standard for assessing scene similarity.  \\
    - The **second image** is the **model's output image**, which you need to evaluate. \\
    
    --- Task: Rate from 0 to 10 \\
    Evaluate whether the **subject from the reference image** has been correctly and faithfully **manipulated or integrated** into the **base image**, in accordance with the **editing instruction**. \\
    
    ---- **Scoring Criteria** \\
    
    - **0:** The manipulation *completely failed* — the reference subject is missing or unrecognizable.  \\
    - **1-3:** The manipulation is *severely incorrect*, showing only vague or partial resemblance to the reference subject.  \\
    - **4-6:** The manipulation is *partially successful* — some recognizable traits of the reference subject appear, but major inconsistencies remain.  \\
    - **7-9:** The manipulation is *mostly accurate*, with strong resemblance to the reference subject and only minor mismatches.  \\
    - **10:** The manipulation is *perfectly successful* — the subject is seamlessly and accurately represented, matching the reference in identity, structure, and style. \\
    
    ---- **Pay special attention to** \\
    
    - **Identity fidelity:** Facial structure, hairstyle, clothing, and other distinctive features should closely match the reference subject.  \\
    - **Pose and spatial alignment:** The manipulated subject should align naturally with the scene's geometry, position, and orientation in the base image.  \\
    - **Expression and attributes:** Facial expressions and physical traits (e.g., age, gender, skin tone) should remain consistent with the reference.  \\
    - **Semantic correctness:** The correct individual or object from the reference should appear exactly where specified by the instruction.  \\
    - **Selective manipulation:** Only the target subject should be replaced or modified — no unintended entities should appear or disappear. \\
    
    ---- **Important Notes** \\
    
    - **Deduct points** for every visible mismatch in identity, shape, or manipulation accuracy.  \\
    - Do *not* consider **background consistency**, **artifact realism**, or **aesthetic appeal** — focus exclusively on the correctness of the subject manipulation itself. \\ 
    - The final score should reflect **how accurately and faithfully the model followed the instruction to manipulate or replace the subject**. \\ 
    - **Scoring should be strict** — assign high scores only when the manipulated subject strongly and consistently matches the reference. \\
    
    **Editing instruction:** $`<instruction>`$ \\
    \end{tcolorbox}
    \caption{Prompt for MF score.}
    \label{fig:prompt_mf_score}
\end{figure*}
\begin{figure*}[t!]
    \centering
    \begin{tcolorbox}[width=\textwidth]
    --- Setting \\
    You are given multiple images: \\
    - The **first image** is the **ground truth (reference)**, serving as a visual standard for assessing scene similarity.  \\
    - The **second image** is the **model's output image**, which you need to evaluate. \\ 
    
    --- Task: Rate from 0 to 10 \\
    Evaluate how well the **non-edited regions** of the **model's output image** remain consistent with the **reference image**, focusing solely on the **unchanged parts of the scene**. \\
    
    This evaluation measures **scene preservation**, not subject replacement accuracy.
    
    ---- **Scoring Criteria** \\
    
    - **0:** The image is *completely inconsistent* — large portions of the original scene are altered, distorted, or missing.  \\
    - **1-3:** The overall scene has *major inconsistencies*, with substantial background changes or unnatural modifications.  \\
    - **4-6:** The main structure of the scene is *partially preserved*, but noticeable distortions, lighting shifts, or color inconsistencies remain.  \\
    - **7-9:** The scene is *mostly consistent*, with only minor local deviations or blending artifacts in non-target regions.  \\
    - **10:** The scene is *perfectly consistent* — all unedited regions are visually identical to the reference, with no perceptible changes. \\
    
    ---- **Pay special attention to** \\
    
    - **Background integrity:** Buildings, furniture, landscape, and other static elements should remain identical to the reference.  \\
    - **Lighting and tone stability:** Global illumination, color temperature, and shading should remain consistent across the entire scene.  \\
    - **Texture and color fidelity:** Non-edited areas should preserve the same texture, hue, and contrast as in the reference.  \\
    - **Spatial structure:** Perspective, geometry, and layout of the environment should be unchanged.  \\
    - **Boundary transitions:** The area surrounding the edited region should blend smoothly into the preserved background without distortion or ghosting. \\
    
    ---- **Important Notes** \\
    
    - Deduct points for any visible deviation, distortion, or inconsistency in regions **unrelated to the instructed edit**.  \\
    - Do *not* evaluate the accuracy of the replaced subject or the realism of the edit itself — focus **only** on the similarity of **unchanged areas**.  \\
    - The score should reflect **how faithfully the original scene was preserved** apart from the intended modification.  \\
    - **Scoring should be strict** — even small but noticeable inconsistencies should lower the score. \\
    \end{tcolorbox}
    \caption{Prompt for BC score.}
    \label{fig:prompt_bc_score}
\end{figure*}

\section{Visualization}
\begin{figure*}[t]
\centering
\includegraphics[width=\linewidth]{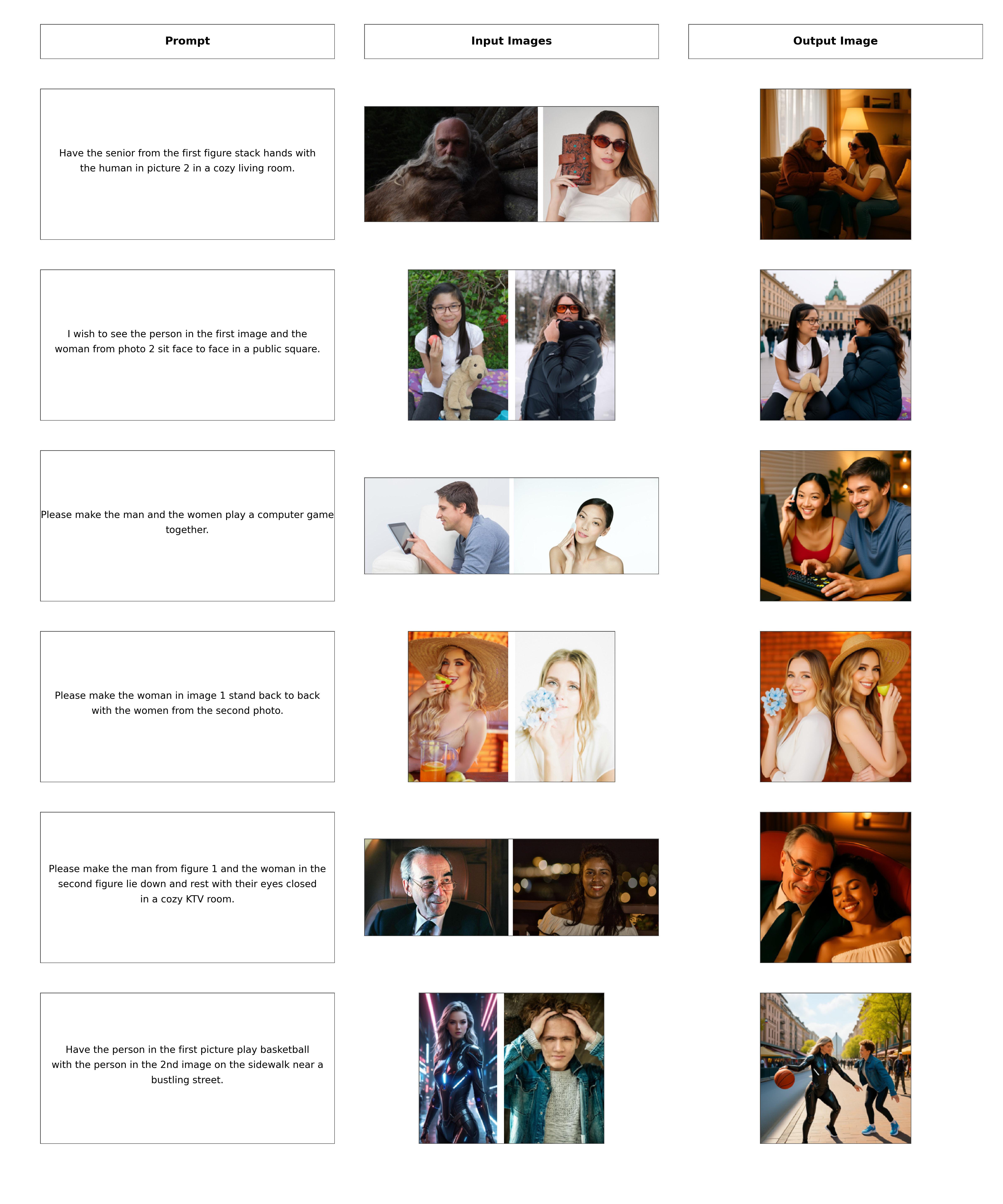}
\caption{Visualization example.}
\label{line_graph1}
\end{figure*}

\begin{figure*}[t]
\centering
\includegraphics[width=\linewidth]{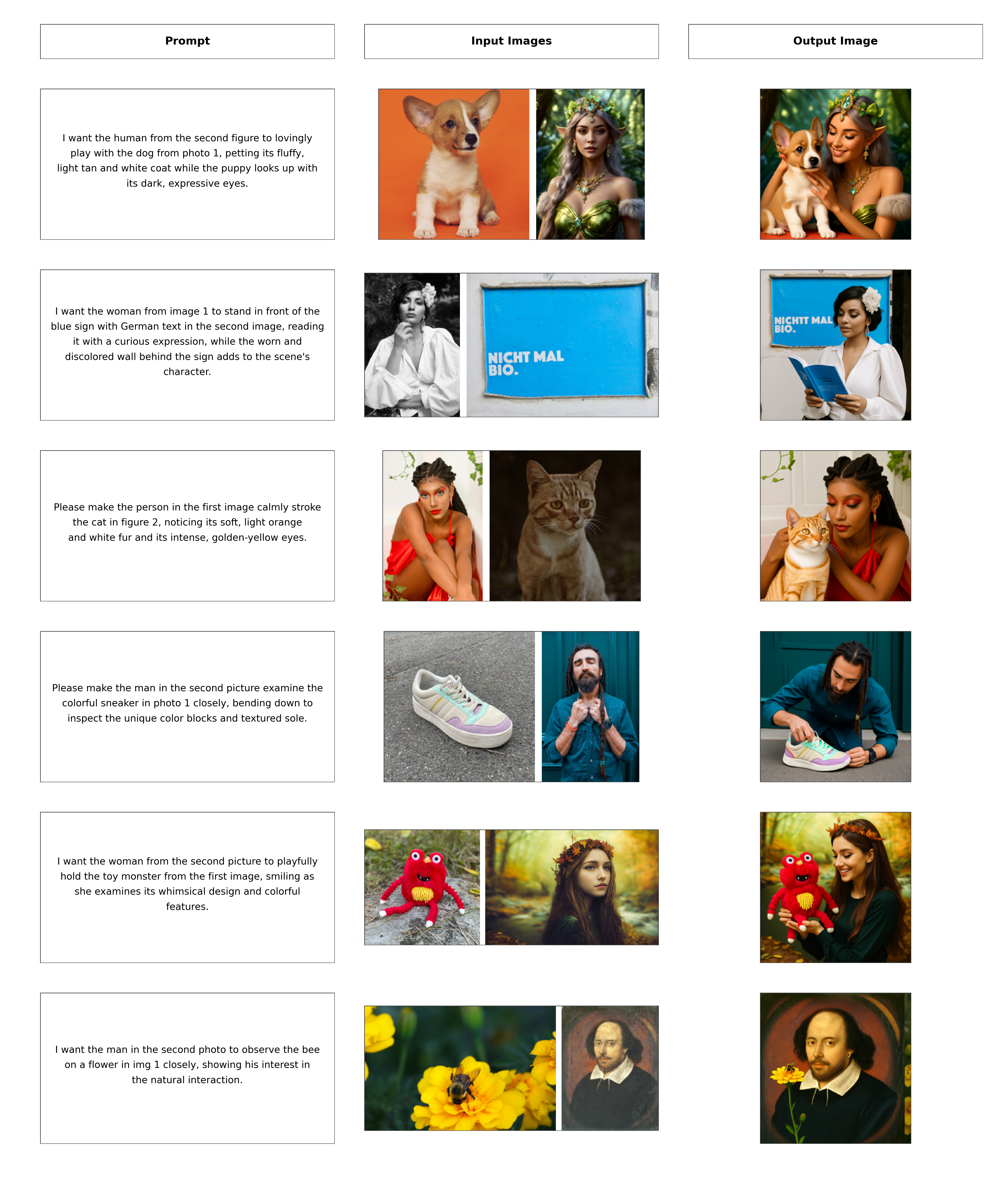}
\caption{Visualization example.}
\label{line_graph2}
\end{figure*}

\begin{figure*}[t]
\centering
\includegraphics[width=\linewidth]{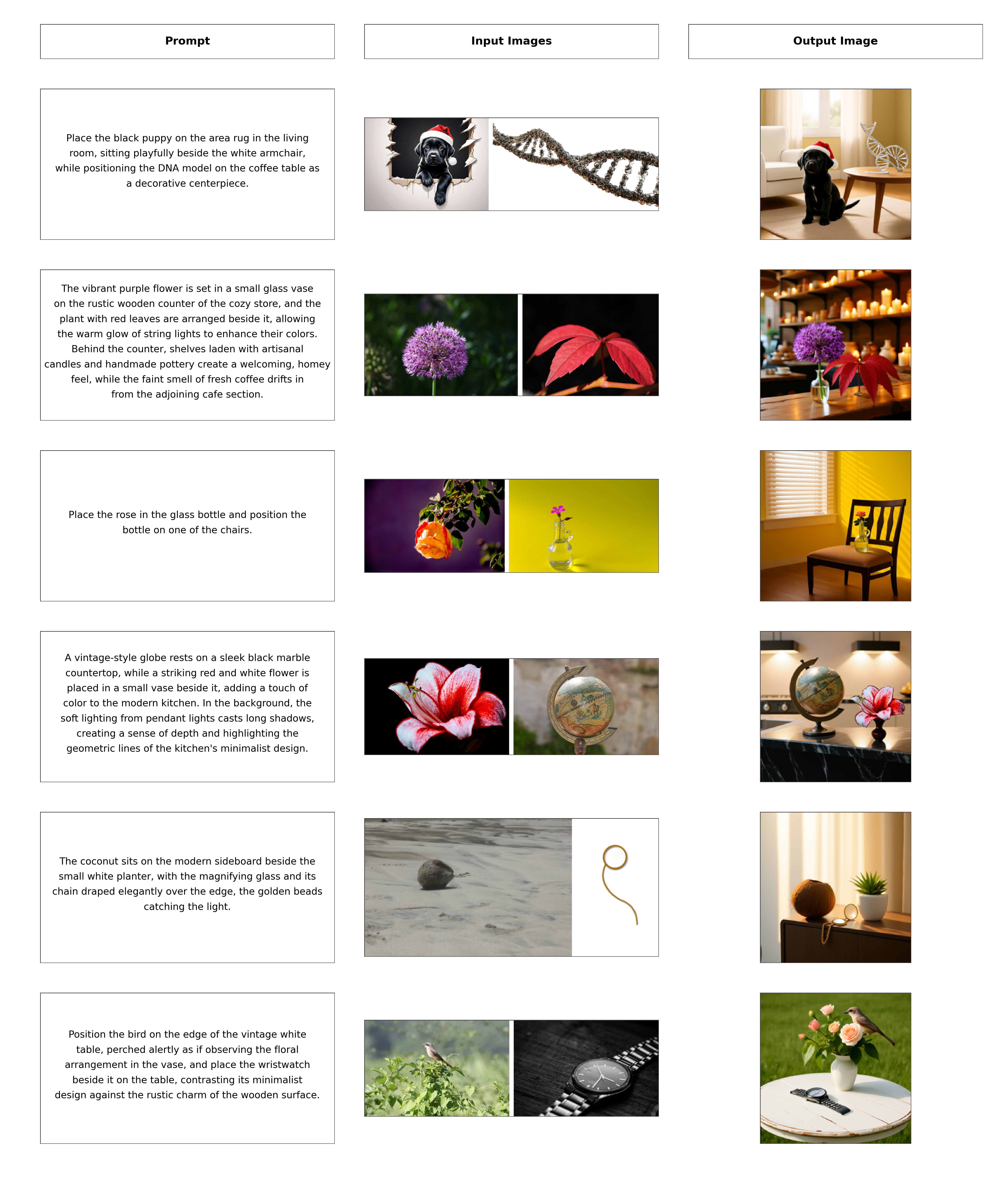}
\caption{Visualization example.}
\label{line_graph3}
\end{figure*}

\begin{figure*}[t]
\centering
\includegraphics[width=\linewidth]{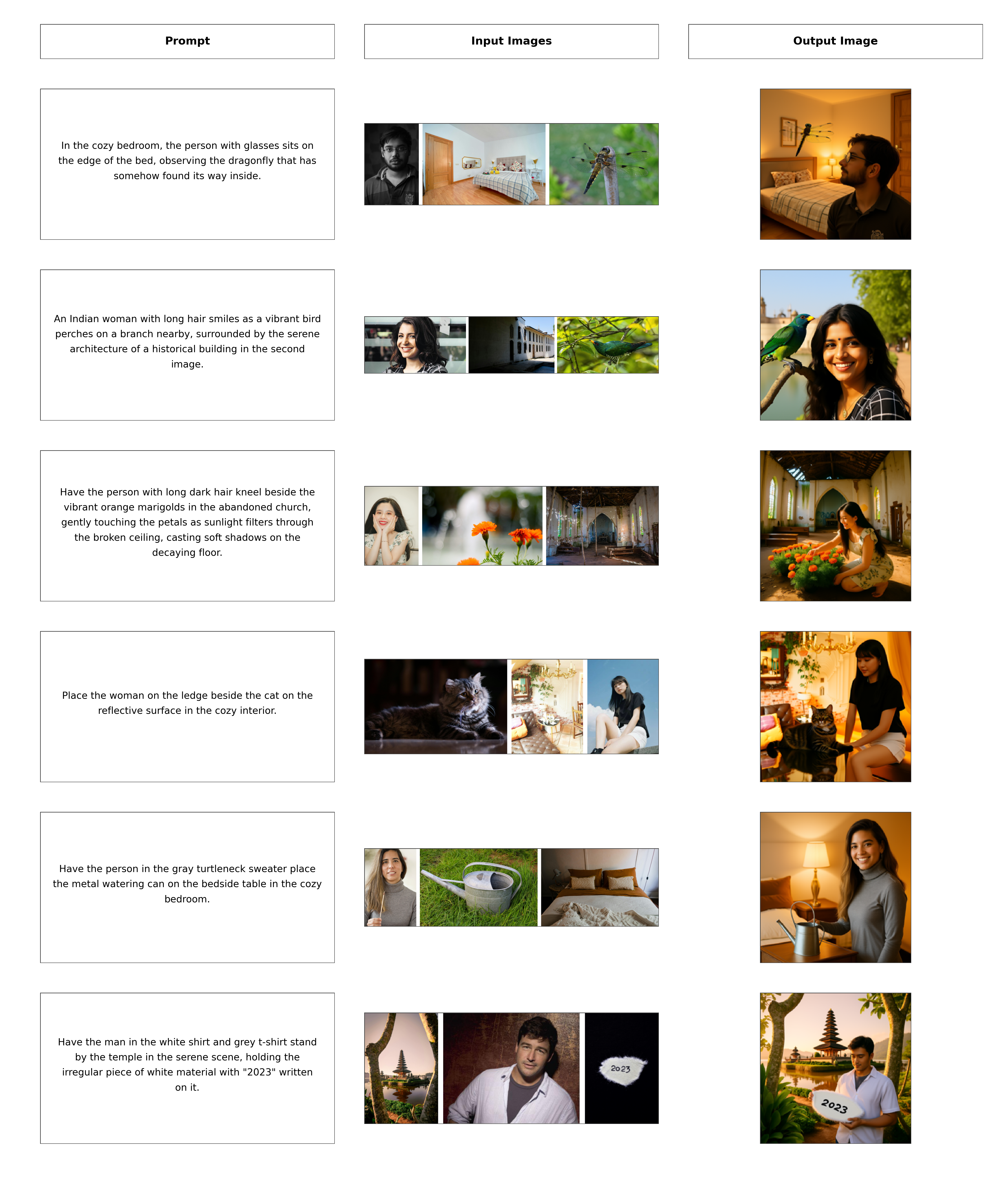}
\caption{Visualization example.}
\label{line_graph4}
\end{figure*}

\begin{figure*}[t]
\centering
\includegraphics[width=\linewidth]{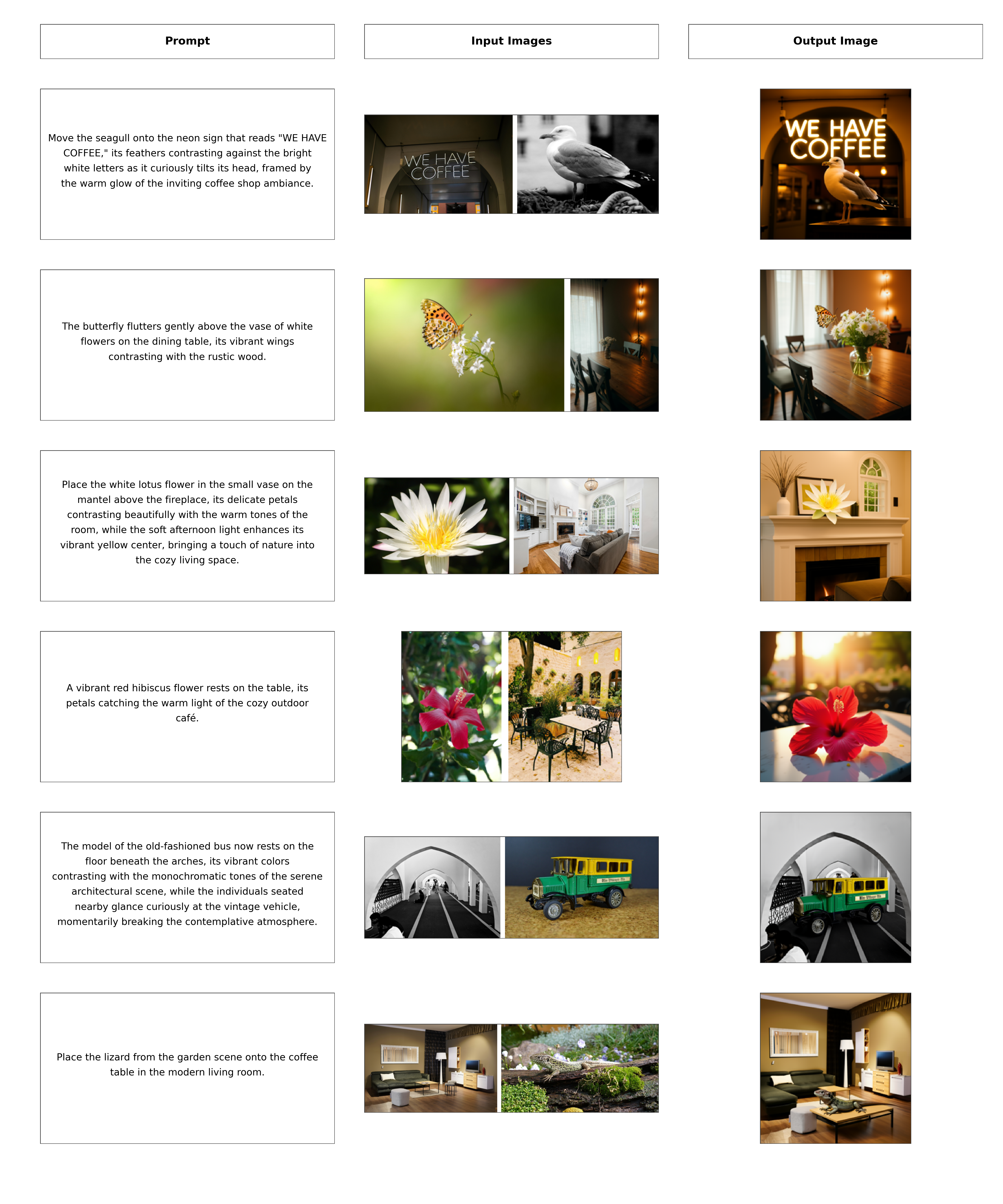}
\caption{Visualization example.}
\label{line_graph5}
\end{figure*}

\begin{figure*}[t]
\centering
\includegraphics[width=\linewidth]{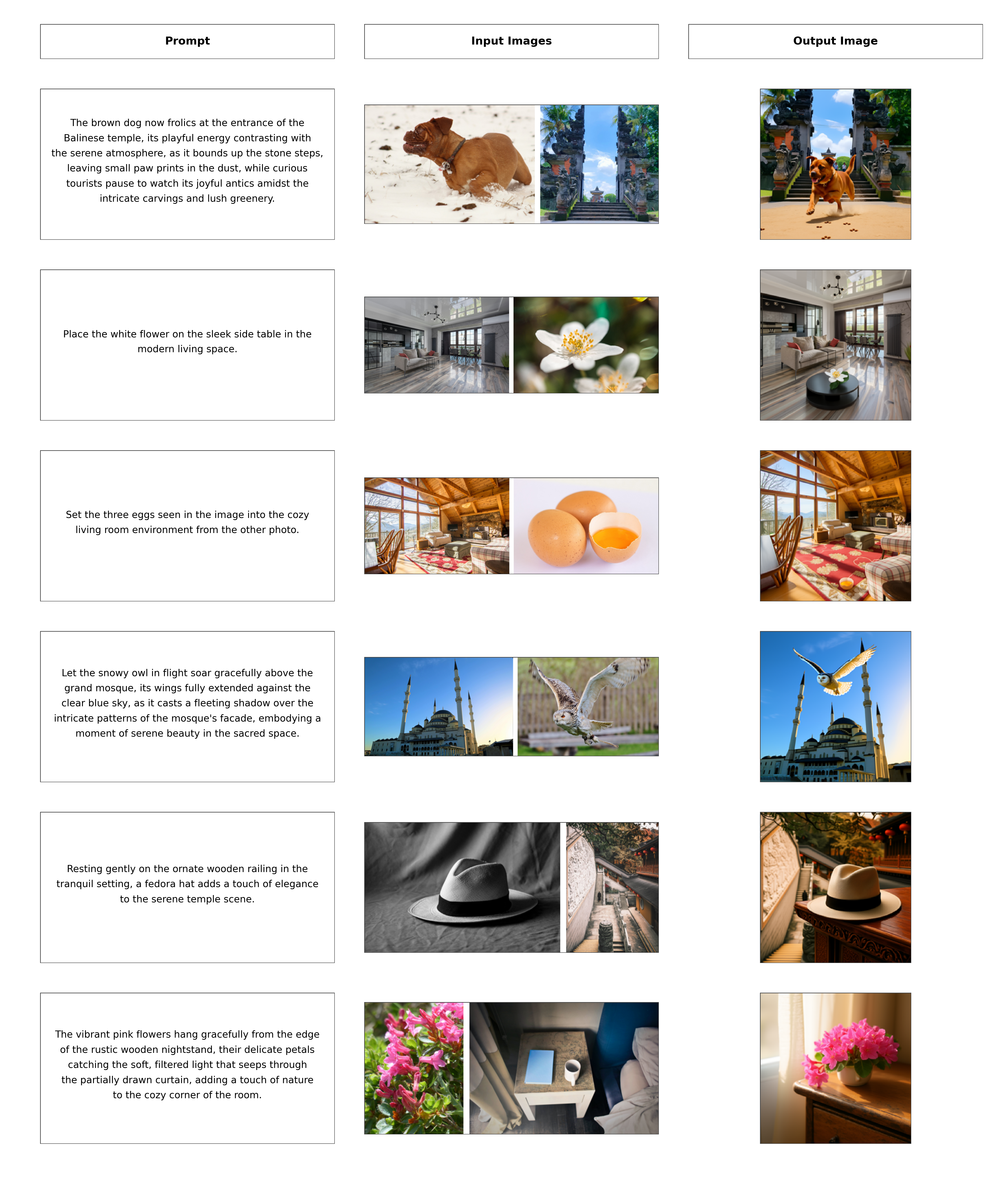}
\caption{Visualization example.}
\label{line_graph6}
\end{figure*}

\begin{figure*}[t]
\centering
\includegraphics[width=\linewidth]{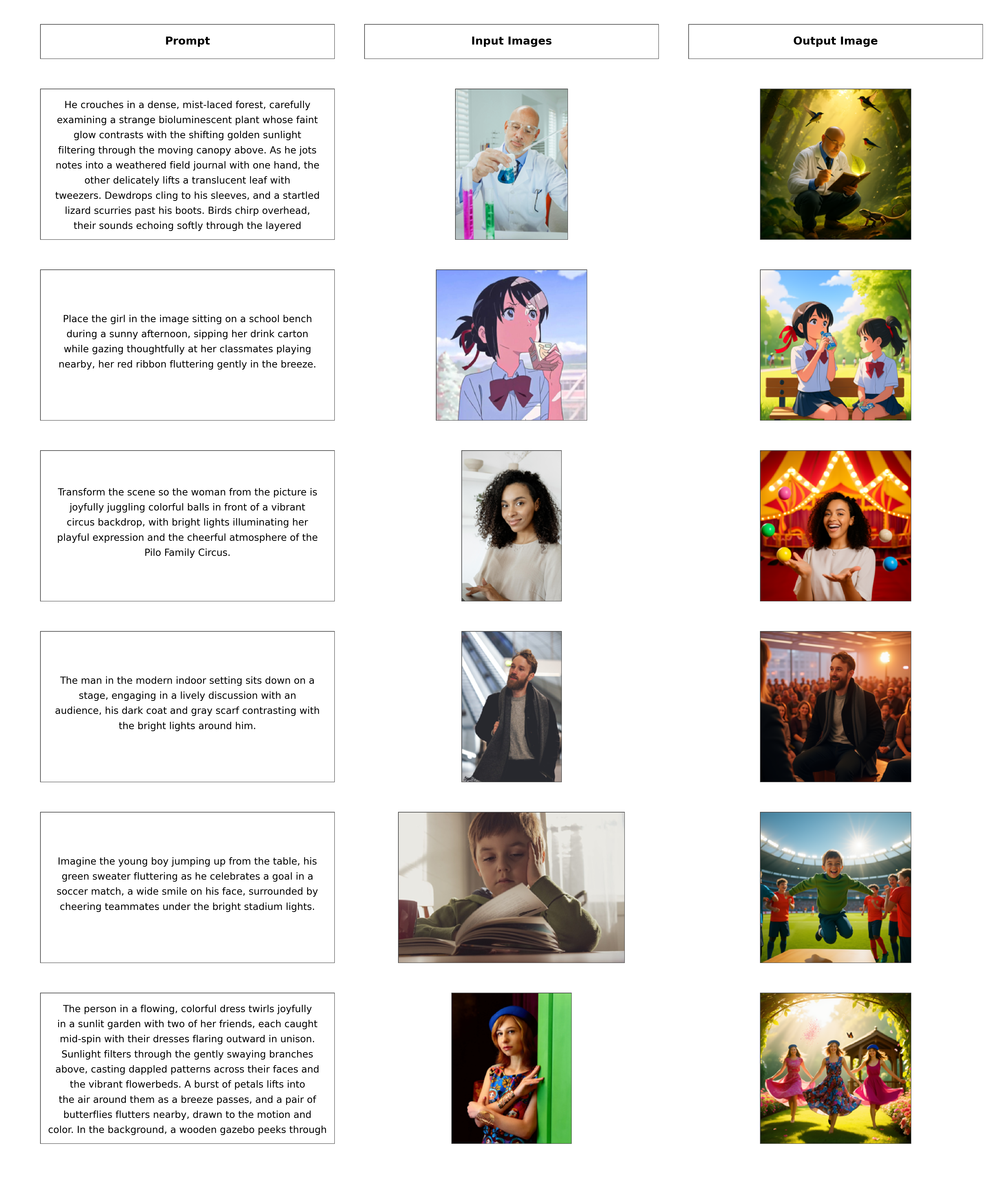}
\caption{Visualization example.}
\label{line_graph7}
\end{figure*}

\begin{figure*}[t]
\centering
\includegraphics[width=\linewidth]{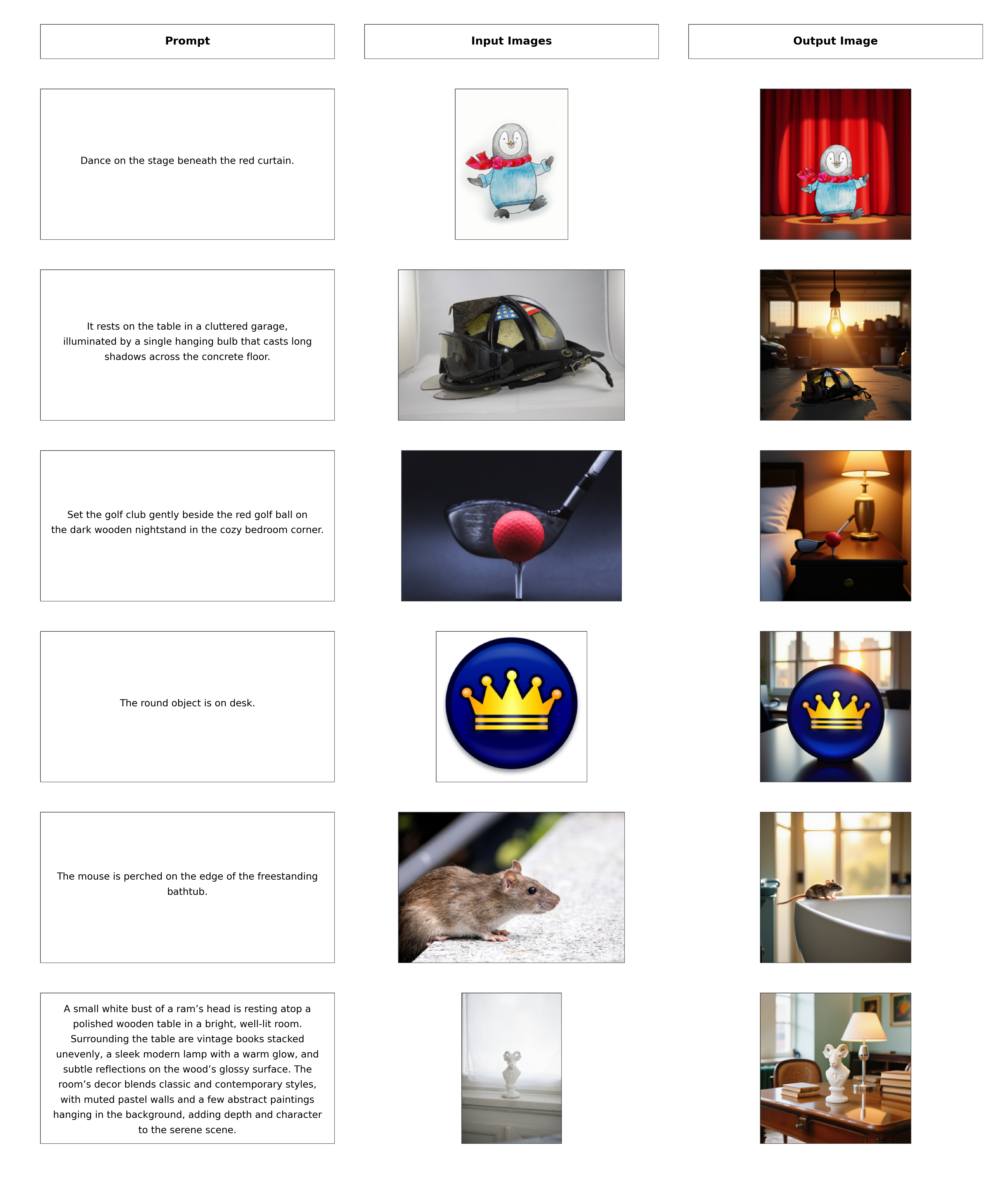}
\caption{Visualization example.}
\label{line_graph8}
\end{figure*}

We further report additional visual results in Fig.~\ref{line_graph1} to Fig.~\ref{line_graph8}. These figures cover a wide range of scenarios.

\end{document}